\pdfoutput=1

\documentclass[11pt]{article}

\usepackage[final]{acl}

\usepackage{times}
\usepackage{latexsym}

\usepackage[T1]{fontenc}

\usepackage[utf8]{inputenc}

\usepackage{microtype}

\usepackage{inconsolata}

\usepackage{graphicx}

\usepackage{booktabs}
\usepackage{amsmath}
\usepackage{tcolorbox}
\usepackage{threeparttable}
\usepackage{algorithm}
\usepackage[noend]{algorithmic}
\usepackage{colortbl}
\usepackage{multirow}
\usepackage{placeins} 
\usepackage{color}
\usepackage{enumitem}

\title{GraphGen: Enhancing Supervised Fine-Tuning for LLMs with Knowledge-Driven Synthetic Data Generation}

\author{
 \textbf{Zihong Chen}\textsuperscript{1} \quad
 \textbf{Wanli Jiang}\textsuperscript{1} \quad
 \textbf{Jinzhe Li}\textsuperscript{1}
\\
 \textbf{Zhonghang Yuan}\textsuperscript{1} \quad
  \textbf{Huanjun Kong}\textsuperscript{1} \quad
  \textbf{Wanli Ouyang}\textsuperscript{1,3}  \quad
 \textbf{Nanqing Dong}\textsuperscript{1,2}\thanks{Corresponding author.}
\\
 \textsuperscript{1}Shanghai Artificial Intelligence Laboratory\quad
 \textsuperscript{2}Shanghai Innovation Institute\\
 \textsuperscript{3}The Chinese University of Hong Kong
}

\begin{document}
\maketitle
\begin{abstract}

Fine-tuning for large language models (LLMs) typically requires substantial amounts of high-quality supervised data, which is both costly and labor-intensive to acquire. While synthetic data generation has emerged as a promising solution, existing approaches frequently suffer from factual inaccuracies, insufficient long-tail coverage, simplistic knowledge structures, and homogenized outputs. To address these challenges, we introduce GraphGen, a knowledge graph-guided framework designed for three key question-answering (QA) scenarios: atomic QA, aggregated QA, and multi-hop QA. It begins by constructing a fine-grained knowledge graph from the source text. It then identifies knowledge gaps in LLMs using the expected calibration error metric, prioritizing the generation of QA pairs that target high-value, long-tail knowledge. Furthermore, GraphGen incorporates multi-hop neighborhood sampling to capture complex relational information and employs style-controlled generation to diversify the resulting QA data. Experimental results on knowledge-intensive tasks under closed-book settings demonstrate that GraphGen outperforms conventional synthetic data methods, offering a more reliable and comprehensive solution to the data scarcity challenge in supervised fine-tuning. The code and data are publicly available at \url{https://github.com/open-sciencelab/GraphGen}.

\end{abstract}

\section{Introduction}
The rapid advancement of large language models (LLMs) has created a growing need for fine-tuning general-purpose models to incorporate new knowledge efficiently. One widely adopted approach is supervised fine-tuning (SFT), which enables LLMs to learn domain-specific information from labeled training data \citep{parthasarathy2024ultimate, lu2024versatune}. While SFT has proven effective in enhancing model knowledge \citep{mecklenburg2024injecting}, its success heavily depends on access to large-scale, high-quality training datasets, which are expensive to curate and require substantial domain expertise.

To mitigate this data bottleneck, researchers have explored LLM-based synthetic data generation \citep{liu2024best}, leveraging LLMs to autonomously generate training samples, such as question-answer (QA) pairs or textual knowledge snippets. Several existing methods \citep{zhang2023selfqaunsupervisedknowledgeguided, maini2024rephrasingwebrecipecompute} attempt to enhance domain adaptation by expanding training resources. However, when applied to knowledge-intensive tasks in closed-book settings, these synthetic data generation pipelines exhibit critical limitations:
\begin{enumerate}[noitemsep, topsep=1pt, leftmargin=*]
\item \textbf{Factual Inaccuracy:} LLMs often introduce factual errors due to their tendency to hallucinate incorrect or non-factual knowledge \citep{long2024llms}, leading to unreliable training data.

\item \textbf{Insufficient Coverage of Long-Tail Knowledge:} Since LLMs are optimized for token prediction, they tend to prioritize generating high-frequency, common knowledge while failing to capture rare, domain-specific information \citep{li2024search}. This results in inadequate coverage of long-tail knowledge, which is crucial for knowledge-intensive applications \citep{kandpal2023large, li2024role}.
\item \textbf{Superficial Knowledge Representation:} Existing synthetic data pipelines generate simplistic QA pairs that do not effectively model complex knowledge structures, such as multi-hop reasoning, where information must be linked across multiple sources to form a coherent answer.
\item \textbf{Homogenization and Overfitting Risks:} Synthetic datasets often suffer from low diversity, with repetitive sentence templates and similar difficulty levels. This lack of variation can lead to overfitting, reducing the generalization ability of fine-tuned models and, in extreme cases, causing catastrophic forgetting or model collapse \citep{shumailov2024ai}.
\end{enumerate}

Recent efforts have attempted to improve synthetic data generation by incorporating Monte Carlo tree search and chain-of-thought reasoning \citep{zhao2024marcoo1openreasoningmodels, wei2022chain}. However, these methods primarily focus on logical problem-solving and do not effectively adapt to knowledge-intensive tasks in closed-book scenarios.

To address these challenges, we propose GraphGen, a knowledge graph (KG)-calibrated data synthesis framework that systematically improves synthetic data quality through structured knowledge guidance. GraphGen is designed to enhance data generation in three key scenarios: \textbf{atomic QA} (covering basic knowledge), \textbf{aggregated QA} (incorporating complex, integrated knowledge), and \textbf{multi-hop QA} (extending to $k$-hop reasoning).

Specifically, we first construct a fine-grained KG from a source corpus. We then compute the expected calibration error (ECE) \citep{guo2017calibration} for each triple in the KG to identify points where the model’s confidence does not align with its actual accuracy, exposing potential \emph{knowledge blind spots}. The framework prioritizes these high-ECE triples for targeted data augmentation. To ensure the contextual coherence of newly generated examples, we employ a $k$-hop neighborhood subgraph sampler with structural constraints. Lastly, we employ style-controlled generation to convert the sampled subgraphs into diverse QA pairs suited for SFT. Our experiments show that GraphGen consistently outperforms five established data synthesis baselines in the aforementioned three scenarios.
In summary, our main contributions are:
\begin{itemize}[noitemsep, topsep=1pt, leftmargin=*]
    \item We propose GraphGen, a KG-based data synthesis framework designed to preserve knowledge associations while addressing limitations in coverage, which is effective for scenarios of atomic QA, aggregated QA, and multi-hop QA.
    \item We develop an ECE-driven module that identifies knowledge blind spots, enabling LLMs to focus on high-value, long-tail data.
    \item Through extensive evaluations, we demonstrate that GraphGen leads to more effective SFT on knowledge-intensive tasks under closed-book conditions than existing state-of-the-art methods.
\end{itemize}


\begin{figure*}[t]
  \includegraphics[width=\linewidth]{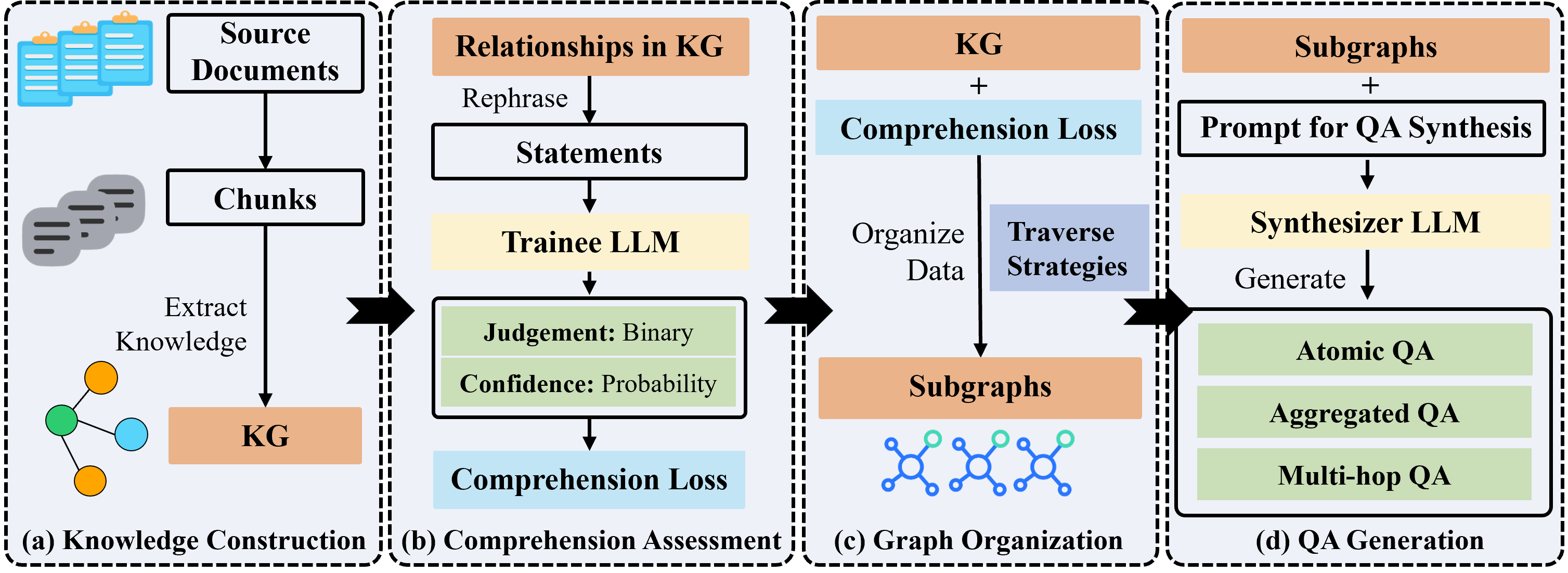}
  \caption {Pipeline of GraphGen. GraphGen optimizes LLM's
 performance by effectively organizing knowledge and identifying the specific data required for training the model. It comprises four core stages: \textbf{Step 1 (a)}: Initially, entities/relationships are extracted to build a KG. \textbf{Step 2 (b)}: Then, the Trainee Model’s understanding of knowledge points is evaluated by judging the correctness of given statements and calculating the comprehension loss accordingly. \textbf{Step 3 (c)}: Then, subgraphs are formed for efficient training. The composition of these subgraphs is controlled using various traversal strategies. \textbf{Step 4 (d)}: Finally, subgraphs are converted into QA pairs for the three scenarios: atomic QA, aggregated QA and multi-hop QA (see Section \ref{sec:graphgen} for details).}
  \label{fig:flow}
\end{figure*}




\section{Related Work}
\label{sec:related_work}
\subsection{Knowledge Graph-based Data Generation}
KGs provide structured representations of domain-specific information, enabling systematic modeling of entities and their relationships. Early KG-based data generation approaches relied on hand-crafted templates \citep{jia2016data, Seyler_2017}, which, despite ensuring syntactic correctness, often produced repetitive and rigid outputs, limiting linguistic diversity and scalability.

To overcome these limitations, learning-based methods leveraging recurrent neural networks with attention mechanisms were introduced to generate fluent questions directly from KG triples \citep{indurthi2017generating, du2017learning}. More recent advancements, such as LFKQG \citep{fei2022lfkqg}, incorporated controlled generation techniques to improve entity coverage while fine-tuning for adaptability. However, ensuring factual consistency and generating high-quality text remain open challenges.

\subsection{LLM-based Data Generation}
LLMs have demonstrated remarkable generalization and reasoning capabilities across natural language tasks \citep{zhang2023selfqaunsupervisedknowledgeguided, maini2024rephrasingwebrecipecompute, köksal2024longformeffectiveinstructiontuning}. In the area of data generation, it has been proposed to generate data using large language models to train smaller models \citep{west2022symbolic}. Unlike KG-driven methods, LLMs can generate diverse, human-like text without reliance on predefined templates \citep{liang2023prompting}. However, they often suffer from limited controllability and hallucination \citep{ji2023towards}, leading to factual inconsistencies.
Also, some methods \citep{zhang2023self} rely on instructions synthesized solely by the model itself and lack a mechanism for structured external knowledge injection. Consequently, they perform poorly on knowledge-intensive tasks where rich domain-specific knowledge is crucial.

Efforts to mitigate these issues include multi-stage refinement pipelines such as Genie \citep{yehudai2024genieachievinghumanparity}, which enhances factual accuracy and coherence. Despite these refinements, ensuring domain-specific precision at scale remains a challenge for standalone LLMs.

\subsection{Combining LLMs and Knowledge Graphs}
To enhance factual consistency, hybrid approaches integrating LLMs with KGs have been explored \citep{guo2024sgsh, zhao2024zero, yang2024syntheticcontinuedpretraining}. These methods leverage KGs to guide text generation, improving reliability while maintaining fluency. However, most focus on general text generation or question answering rather than synthetic data generation for SFT.



\section{Problem Setup} \label{sec:promblem_setup}
\paragraph{Synthesizing Data from Raw Corpora}
We focus on approaches that transforms raw text corpora $D _{\text{source}}$ into structured synthetic data $D_{\text{synth}}$. To achieve this, We propose a synthesis algorithm $A_{\text{synth}}$ to generate data. Specifically, we utilize an algorithm $A_{\text{organize}}$ that performs constrained graph traversal to extract subgraphs. The systematic workflow can be represented as follows:
\begin{equation}
  \label{eq:framework}
  A_{\text{synth}}: D_{\text{source}} \xrightarrow{A_{\text{extract}}} KG \xrightarrow{A_{\text{organize}}} D_{\text{synth}}
\end{equation}


\paragraph{Evaluating the Quality of Synthetic Data}
The quality assessment of synthetic data necessitates both intrinsic quantitative analysis and validation through downstream tasks. We establish a set of multi-dimensional metrics  $Met = \{\text{Metric}\}_{i=1}^n$ for data quality estimation. Additionally, we construct unbiased evaluation datasets $D_{\text{eval}}$ to ensure task-specific validity. The performance on knowledge-intensive QA tasks under closed-book scenarios serves as critical evidence for testing whether the post-SFT model $M_f$ has effectively acquired the knowledge in its parameters. The composite quality metric is formalized as:
\begin{equation}
  \label{eq:metric}
  Q_{D_{\text{synth}}} \propto (s(Met, D_{\text{synth}}), s(D_{\text{eval}},M_f))
\end{equation}
\noindent where $s(Met, D_{\text{synth}})$ denotes the score of $D_{\text{synth}}$ on the metrics, and $s(D_{\text{eval}},M_f)$ indicates the performance of $M_f$ on $D_{\text{eval}}$.

\section{Method} \label{sec:graphgen}
\begin{figure}[t]
  \includegraphics[width=\linewidth]{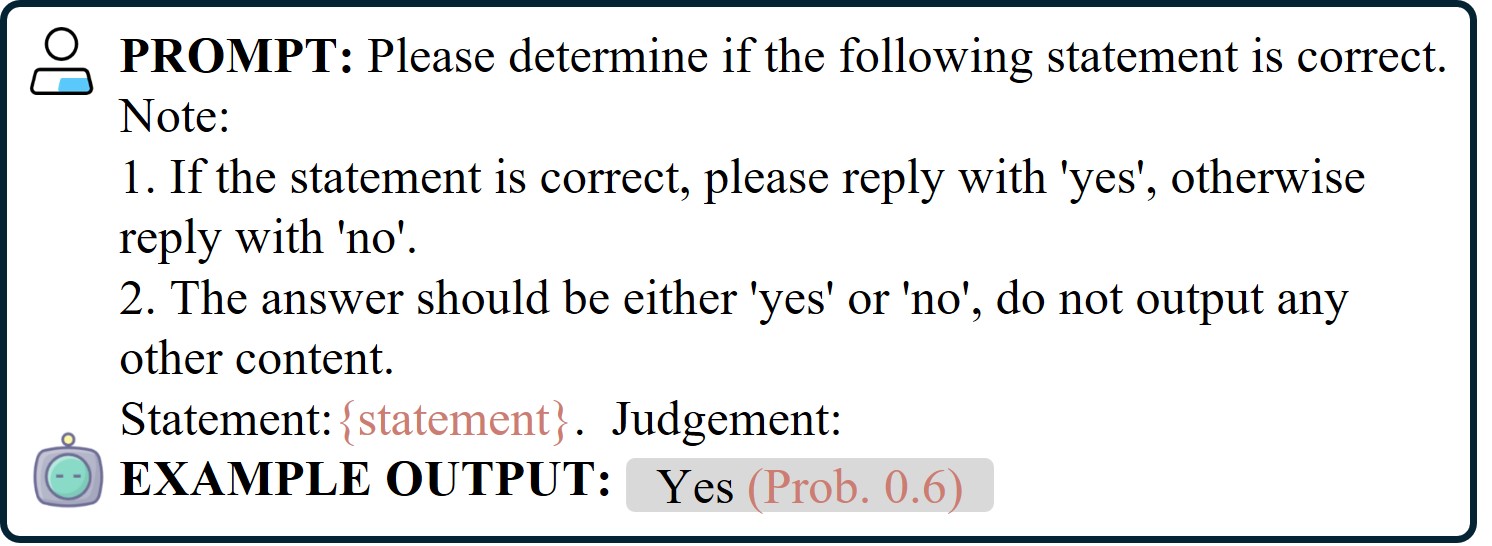}
  \caption{\label{prompt:statement_assessment}Prompt for comprehension assessment. Through binary yes/no questions, we capture precise semantic information for confidence modeling.}
\end{figure}

In this section, we present GraphGen, a data synthesis framework, as illustrated in Figure~\ref{fig:flow}. 
GraphGen is designed to generate data across three scenarios: atomic QA, aggregated QA, and multi-hop QA. From a knowledge organization perspective, these scenarios exemplify the most representative knowledge-intensive tasks in the context of closed-book QA. The framework comprises a four-step workflow involving two interdependent LLMs: the Synthesizer Model ($M_{\text{synth}}$) and the Trainee Model ($M_{\text{train}}$).
$M_{\text{synth}}$ possesses advanced general capabilities, as it is tasked with knowledge extraction and rephrasing.  In contrast, $M_{\text{train}}$ serves as the target model that we aim to enhance in order to integrate additional knowledge. Detailed information regarding the prompt templates utilized in GraphGen, intermediate examples, and implementation details can be found in Appendix~\ref{appendix:graphgen_details}.
\paragraph{STEP 1: Knowledge Construction}
Raw documents are segmented into smaller, semantically coherent fragments through context-aware chunking. Subsequently, $M_{\text{synth}}$ are employed to extract various entities and their relationships from these fragments. The types of entities to be extracted are predefined, with general categories including dates, locations, and events, while domain-specific categories encompass concepts such as genes.  
During the extraction process, if the same entity or relationship appears in multiple fragments, their descriptions will be automatically combined together. Finally, cross-fragment entities and relationships are aggregated into a KG $G=(E, R)$. 
The combination of LLMs and KGs interrelates the \textit{atomic} knowledge, addressesing challenges like long-text processing, format noise, and scattered knowledge distribution, while also ensuring a low rate of hallucination in the generated content \citep{ibrahim2024survey, gillani2024knowledge}.
The specific implementation of STEP 1 is modified from the previous work \citep{guo2024lightragsimplefastretrievalaugmented, huixiangdou2}.
\paragraph{STEP 2: Comprehension Assessment}
We propose a method to assess whether the Trainee Model $M_{\text{train}}$ have fully comprehended a knowledge point from the KG.
For each edge in the KG, its description can be considered as a declarative statement $R_i$, which represents a knowledge point $K_i$ that is unequivocally true, with a real-world probability of 1  (\emph{i.e}., $P(R_i \ \text{is} \ \text{true})=1$).
To evaluate $M_{\text{train}}$'s understanding capabilities of these statements, we first generate multiple paraphrased statements $R_{i1}, R_{i2}, \dots, R_{in}$ and their negations $\neg R_{i1}, \neg R_{i2}, \dots, \neg R_{in}$ using $M_{\text{synth}}$. 
Following the principle of ECE, a model is considered well-calibrated if its predicted confidence scores (\emph{i.e}., softmax probabilities) align with real-world probabilities of correctness. For LLMs, true understanding of a concept is achieved only when the model’s confidence estimates match the actual likelihood of correctness in the real world. Therefore, we use a prompt (see Figure~\ref{prompt:statement_assessment}) to elicit $M_{\text{train}}$'s confidence in a single paraphrased statement. Then, by averaging the confidence scores from the $n$ positive and $n$ negative samples of $R_i$, $M_{\text{train}}$'s confidence in $R_i$ is quantified via the following formula:
\begin{equation}
  \label{eq:confidence}
  \begin{aligned}
  C_{R_i}=\frac{1}{2n} (\sum_{j=1}^n P(t|R_{ij}) + \sum_{j=1}^{n} P(f|\neg R_{ij}))
  \end{aligned}
\end{equation}
\noindent where $P(t|R_{ij})$ is the probability of the next token being ``yes'' given a true statement and $P(f|\neg R_{ij})$ denotes the probability of ``no'' in response to a false statement.










We further define a comprehension loss by calculating the cross-entropy between the true distribution and the predicted distribution:
\begin{equation}
  \label{eq:conprehension_loss}
  \begin{aligned}
    \text{Loss}_{C_{R_i}} = & -\frac{1}{2n}\sum_{j=1}^n log(P(t|R_{ij})) \\
            & - \frac{1}{2n}\sum_{j=1}^n log(P(f|\neg R_{ij}))
  \end{aligned}
\end{equation}


\noindent which measures the gap between the LLM's current understanding and complete mastery of the knowledge point. By assessing the comprehension loss of $M_{\text{train}}$, we can systematically evaluate whether further training with these knowledge points is needed. 
\paragraph{STEP 3: Graph Organization}
Subgraphs are the minimal QA pair generation units.
We perform $k$-hop subgraph extraction for effective graph organization, as detailed in Algorithm ~\ref{alg:khop}. To regulate the composition of these subgraphs, we implement several traverse strategies.
The depth strategy controls the $k$-hop depth, ensuring the subgraph spans a predefined number of hops from the start edge. 
For each candidate subgraph,  we compute the premise length (denoted as $pre\_length$), defined as the total number of tokens in the descriptions of entities and relationships within it. The length strategy enforces an upper bound on $pre\_length$ to maintain a balanced data distribution.
When expanding the subgraph, we adopt a selection strategy with three options:
\begin{enumerate}[noitemsep, topsep=1pt, leftmargin=*]
\item max\_loss: Select edges with higher loss values, indicating greater uncertainty or potential information gain.
\item min\_loss: Select edges with lower loss values, representing more confident or stable relations.
\item random: Select edges uniformly at random.
\end{enumerate}
These strategies collectively balance subgraph complexity, relevance, and computational tractability.
\paragraph{STEP 4: QA Generation}
After extracting a subgraph, we can create three types of QA pairs based on its intended use. For the atomic QA scenario, the subgraph should consist of a single node or edge, allowing $M_{\text{synth}}$ to generate a QA pair representing basic knowledge.
To analyze, summarize, or compare related information involving a set of entities and relationships within a subgraph, we prompt $M_{\text{synth}}$ to organize and rephrase the data into a coherent text(the answer). Then  we use $M_{\text{synth}}$ to generate its corresponding question. 
For multi-hop QAs, we first clarify the relationships between entities and then instruct $M_{\text{synth}}$ to produce a QA pair that requires multi-step reasoning.

\begin{algorithm}[t]
\caption{$K$-hop Subgraph Extraction}
\label{alg:khop}
\renewcommand{\algorithmicrequire}{\textbf{Input:}}
\renewcommand{\algorithmicensure}{\textbf{Output:}}
\newcommand{\BREAK}{\STATE \textbf{break}}
\begin{algorithmic}[1]
\REQUIRE
    Graph $G$, edge $R_i = (E_{\text{src}}, E_{\text{tgt}})$, graph organization strategies $S$
\ENSURE
    Subgraph $G'$

    \STATE $G' \gets \{R_i\}$ 
    \STATE $C \gets \textsc{GetAdjacentEdges}(G, \{E_{\text{src}}, E_{\text{tgt}}\})$

\WHILE{$C \neq \emptyset$}
    \STATE Select $e$ from $C$ according to $S$
    \STATE $G' \gets G' \cup \{e\}$
    \STATE $C \gets C \setminus \{e\}$

    \IF{$\textsc{MeetsConstraints}(G')$}
        \BREAK
    \ENDIF

    \FOR{$v \in \textsc{GetEndpoints}(e)$}
        \STATE $C \gets C \cup \textsc{GetAdjacentEdges}(G, v)$
    \ENDFOR
\ENDWHILE
\RETURN $G'$ 
\end{algorithmic}
\end{algorithm}

\section{Experiments}
\label{sec:experiments}
\begin{figure*}[t]
  \centering
  \includegraphics[width=\linewidth]{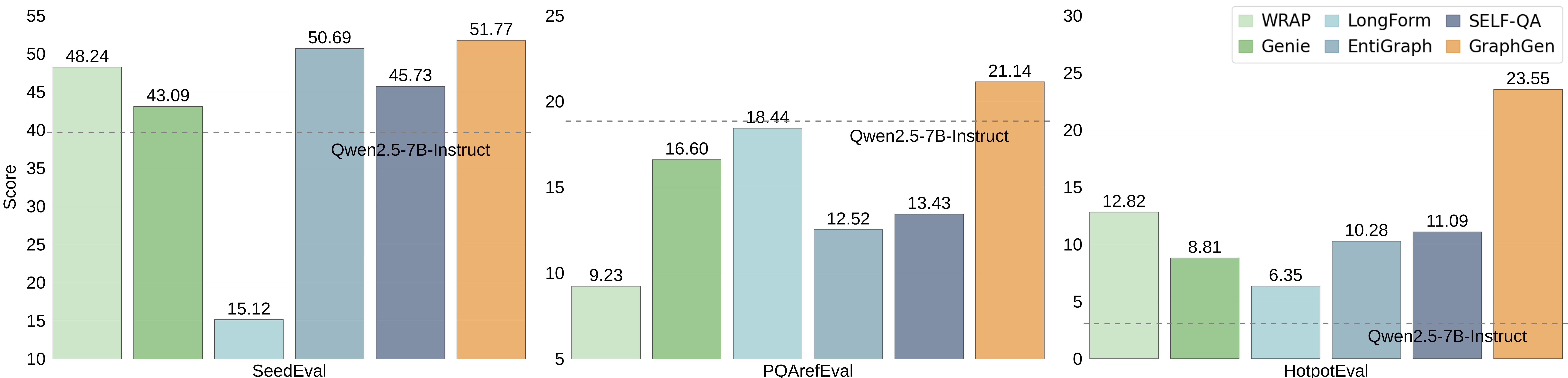}
  \caption{\label{exp:eval}Performance comparison on knowledge-intensive evaluation datasets. We use data generated through various methods to optimize Qwen2.5-7B-Instruct. We use ROUGE-F as the metric. The baseline methods exhibit varying performance across the three datasets, while GraphGen consistently achieves optimal results.}
\end{figure*}

\begin{table*}[t]
  \centering
  \small
  \begin{tabular}{l l l l r r r r r r}
    \toprule
    \textbf{Dataset} & \textbf{Domain} & \textbf{Scenario} & \textbf{Language} & \multicolumn{2}{c}{\textbf{\#Samples}} & \multicolumn{2}{c}{\textbf{Avg \#Tokens}} & \multicolumn{2}{c}{\textbf{Max \#Tokens}} \\
    \cmidrule(r){5-6}
    \cmidrule(r){7-8}
    \cmidrule(r){9-10}
    & & & &  Corpus & Test & Corpus & Test  & Corpus & Test \\
    \midrule
    \textit{SeedEval} & Agricultural & Atomic QA & English \& Chinese & 30,578 & 582 & 328 & 48 & 908 & 194\\
    \textit{PQArefEval} & Medical & Aggregated QA & English & 58,078 & 5,815 & 357 & 518 & 1,837 & 1,680\\
    \textit{HotpotEval} & General & Multi-Hop QA  & English & 73,642 & 7,405 & 135 & 25 & 1,985 & 82 \\

    \bottomrule
  \end{tabular}
  \caption{\label{tab:dataset}Description of datasets employed for experiments. For calculating the token count, the tokenizer used is from Qwen2.5 series \citep{qwen2025qwen25technicalreport}. The corpus is employed for graph construction and data synthesis, while the test set is utilized to evaluate the performance of the Post-SFT model trained with the synthesized data.}
\end{table*}

\subsection{Experimental Setup}

\paragraph{Domain Corpus and Evaluation Datasets}

To target knowledge-intensive tasks in closed-book QA, we utilized three datasets, each aligned with a critical scenario. We adapted the domain-specific dataset SeedEval from SeedBench~\cite{yingseedbench2025}, a benchmark related to seed knowledge (agriculture), which cover one-shot and zero-shot scenarios.
Additionally, we adapted the \textit{PQArefEval} dataset from \textit{PQAref} \cite{bašaragin2024knowthatteachinggenerative}, which is domain-specific and centers on medicine, constructing it for aggregated QA applications. Furthermore, we created \textit{HotpotEval}, an adaptation of \textit{HotpotQA} \cite{yang2018hotpotqadatasetdiverseexplainable}, intended for multi-hop QA tasks. Each dataset comprises two components: the QA test set ($D_{\text{eval}}$) and the corresponding source texts ($D_{\text{source}}$). See Appendix~\ref{appendix:dataset_details} for the source and details of the dataset. 

\paragraph{Quality Evaluation Metrics}
We employ a set of natural language metrics \cite{cao2024instructionmininginstructiondata} to evaluate the quality of generated text. 
Details are provided in Appendix~\ref{appendix:metric_evaluation_details}.
Since most of these metrics are better suited for evaluating complete sentences than brief responses, we compared the aggregated QAs produced by GraphGen with those from baseline methods. The reward score averages scores from two reward models, labeled as \textit{Ind} and \textit{Deb}. The unieval score comprises three evaluation components from the UniEval model, denoted as \textit{Nat}, \textit{Coh}, and \textit{Und}. 
\paragraph{Baselines}
We modified the code for WRAP \citep{maini2024rephrasingwebrecipecompute}, Genie \citep{yehudai2024genieachievinghumanparity}, LongForm \citep{köksal2024longformeffectiveinstructiontuning}, EntiGraph \citep{yang2024syntheticcontinuedpretraining}, and SELF-QA \citep{zhang2023selfqaunsupervisedknowledgeguided} to accommodate our data synthesis needs, using them as the baselines for this study. See Appendix~\ref{appendix:baseline_details} for details of the baselines.
\paragraph{Implementation Details}
In this study, we specified $M_{\text{train}}$ to be Qwen2.5-7B-Instruct
and $M_{\text{synth}}$ to be Qwen2.5-72B-Instruct
. Two models are representative open-source LLM with robust performance and affordable computational cost.
Considering the characteristics of the tasks associated with the three datasets, and to thoroughly validate our methodology, GraphGen generates atomic, aggregated, and multi-hop QA pairs for dataset \textit{SeedEval}, \textit{PQArefEval}, and \textit{HotpotEval}, respectively. Additional setups can be found in Appendix~\ref{appendix:additional_setups}.

\subsection{Performance Comparison}

\paragraph{Results on Quality Evaluation Metrics}

\begin{table*}
  \centering
  \small
  \begin{tabular}{l r r r r r r r r r}
    \toprule
    \textbf{Method} & \textbf{\#Samples} & \textbf{Avg \#Tokens} & \multicolumn{6}{c}{\textbf{Results}} & \textbf{Avg Score}\\
    \cmidrule(r){4-9}
    & & & \textbf{MTLD} & \multicolumn{3}{c}{\textbf{Uni}} & \multicolumn{2}{c}{\textbf{Rew}} \\
    \cmidrule(r){5-7}
    \cmidrule(r){8-9}
    & & & & Nat & Coh & Und & Ind & Deb  \\
    \midrule
    WRAP       & 476,626 & 32.4 & 13.4 & 91.2 & 87.1 & 91.6 & 44.0 & 4.0 & 42.5  \\
    Genie      & 56,938 & 83.8 & 40.2 & \cellcolor{black!10}91.7 & \cellcolor{black!10}94.7 & \cellcolor{black!10}92.7 & 64.1 & \cellcolor{black!10}44.0 & 62.4  \\
    LongForm   & 57,854 & 357.1 & \cellcolor{black!10}47.6 & 85.7 & 93.7 & 87.3 & \cellcolor{black!10}84.2 & \cellcolor{black!20}82.5 & \cellcolor{black!10}73.3  \\
    EntiGraph & 532,971 & 47.4 & 30.1 & \cellcolor{black!20}92.6 & 93.3 & \cellcolor{black!20}93.1 & 56.3 & 28.8 & 55.2 \\
    SELF-QA    & 561,798 & 83.4 & 34.7 & 91.3 & 92.8 & 92.3 & 59.5 & 39.3 & 58.8 \\
    GraphGen & 54,287 & 657.9 & \cellcolor{black!20}75.8 & 87.8 & \cellcolor{black!20}95.7 & 90.4 & \cellcolor{black!20}85.0 & 31.8 & \cellcolor{black!20}75.2  \\
    \bottomrule
  \end{tabular}
  \caption{\label{exp:metrics}Comparison results with other data synthesis methods on data quality evaluation metrics. The results indicate that the quality of data generated by GraphGen is comparatively high. The scores presented represent the average values derived from the generated datasets across the evaluated metrics. The top two performers in each column are highlighted.}
\end{table*}


We demonstrate that the metrics can be used to intuitively measure data quality. We compare the aggregated responses generated by GraphGen for the aggregated QA scenario with those from baseline methods. As shown in Table~\ref{exp:metrics}, GraphGen outperforms the best baseline method by 1.9 points. Notably, on the MTLD metric for lexical diversity, GraphGen achieves 75.8, surpassing the best baseline method by 28.2 points. GraphGen excels in the MTLD metric due to its capability to aggregate cross-document knowledge, generating a significantly larger number of tokens compared to other methods that yield shorter QA responses. We note that graph-based methods lead the Uni-Score metrics, indicating that data generated through graph structures—particularly those illustrating multiple entity relationships—align more closely with everyday QA interactions. Notably, since LongForm directly uses $D_{\text{source}}$ as the answer in a QA pair, it reflects the quality of $D_{\text{source}}$. Possibly influenced by its training corpus, the Deb metric exhibit a distinct bias towards the original text, which may not be suitable for more chaotic $D_{\text{source}}$. 
\paragraph{Results on Evaluation Datasets}
We conducted SFT on $M_{\text{train}}$ using generated data and evaluated $M_f$ on corresponding test sets, with the results shown in Figure~\ref{exp:eval}.
Data generated by GraphGen brought the greatest performance improvement to the base model. 
 On SeedEval, PQArefEval, and HotpotEval, GraphGen exceeds the best baselines by 1.08, 2.7, and 4.73 points, respectively.
Notably, we observed that on the \textit{PQArefEval} dataset, the performance of baseline methods after training  was inferior to their pre-training performance, which is counterintuitive. We hypothesize that this decline is due to the limitation of baseline methods, which use solely a single text segment for generating QA pairs when handling the aggregated QA task. Consequently, these models may lose their ability to form cross-document associations, negatively impacting their performance on tasks that require multiple references. In contrast, $M_f$ using data generated by GraphGen successfully addresses this challenge. Moreover, GraphGen's performance on the multi-hop QA scenario is particularly notable. This indicates that the knowledge associations derived from subgraphs enhance the multi-hop reasoning capabilities of the Post-SFT model, rather than merely enabling the acquisition of superficial knowledge. 
The variability in performance across different datasets stems from the adaptability issues of baseline methods regarding domain or stylistic differences. For instance, LongForm directly uses the original text from the corpus as answers and generates corresponding questions. For shorter corpora, such as SeedEval and HotpotEval, the synthesis models can effectively adhere to instructions and produce appropriate questions. However, in longer corpora like PQArefEval, the quality of generated questions often declines, leading to suboptimal training outcomes.

It is noteworthy that at this stage, we only used knowledge-related data, without mixing general instruction-following data. The purpose was to highlight the role of generated data in injecting new knowledge. 
In addition, to alleviate concerns about overfitting caused by synthetic data, we mixed the generated data with 100,000 general instruction-following data and conducted tests on a broader evaluation dataset. See Appendix~\ref{appendix:additional_experimental_results} for test results.


\paragraph{Sensitivity of $M_{\text{train}}$}
To further verify that the observed performance improvements are primarily attributed to the quality of the synthesized data rather than the specific characteristics of $M_{\text{train}}$, we conducted additional experiments using two representative open-source models: Meta-Llama-3.1-8B-Instruct and MiniCPM3-4B. The selection of these models is motivated by their distinct architecture types and parameter scales, allowing us to test the robustness and generality of our method across varying LLM structures. The results of two models (detailed in Appendix~\ref{appendix:add_main_experiments}) exhibit trends consistent with our main experimental results. Specifically, GraphGen consistently achieves superior results compared to baseline methods in all three evaluation datasets. These findings strongly suggest that the effectiveness of our approach is independent of specific LLM architectures or parameter sizes, reinforcing the conclusion that the quality and structure of the synthesized data are the primary contributors to performance enhancement.

\subsection{Analysis of Scaling Law}

\begin{figure}[t]
\centering
  \includegraphics[width=\linewidth]{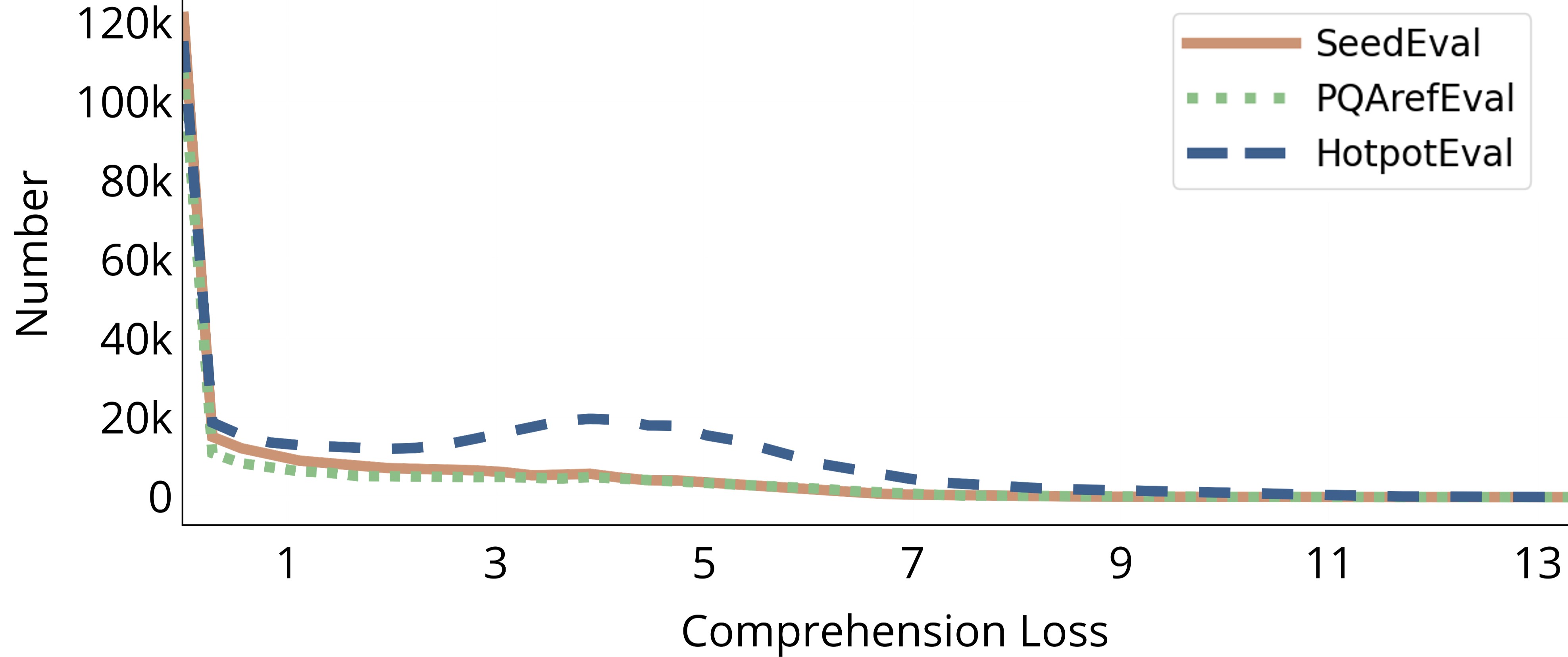}
  \caption{\label{exp:loss_distribution}Distribution of comprehension loss for the Trainee Model. The model's comprehension loss is relatively low for the vast majority of data, which indicates most of the data generated by the Synthesizer Model has already been mastered by the Trainee Model.} 
\end{figure}

\begin{figure}[t]
\centering
  \includegraphics[width=\linewidth]{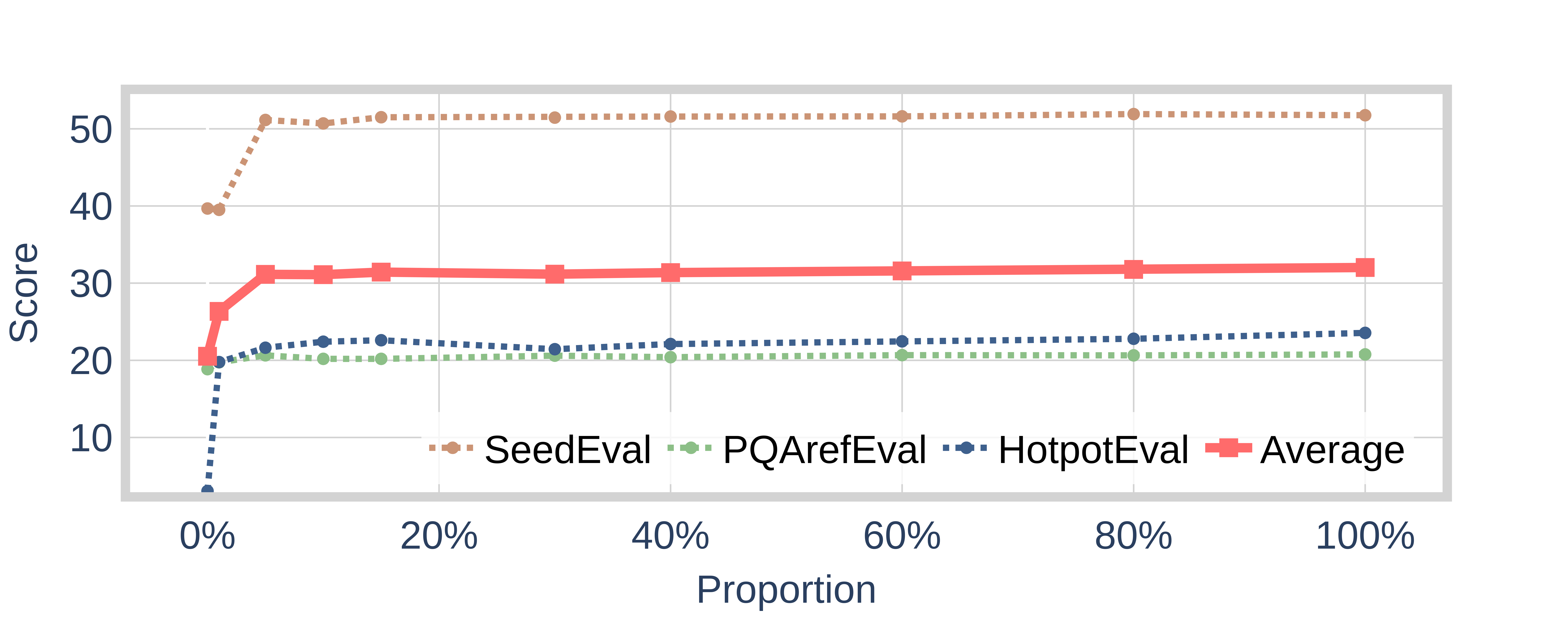}
  \caption{\label{exp:scaling} Performance comparison conducted with varying proportions of training data. The proportions are arranged in descending order based on loss. ``Average'' represents the mean score across three datasets. As the amount of training data increases, we observe a noticeable and consistent upward trend in the results.}
\end{figure}


The scaling law of LLMs shows better model performance with more training data \citep{kaplan2020scalinglawsneurallanguage}. In this study, we obtained the comprehension loss for each knowledge point used in training the model.  Through statistical analysis of the $\text{Loss}_C$ for all knowledge points in the KG, we observed that the distribution of $\text{Loss}_C$ is highly skewed, as illustrated in Figure~\ref{exp:loss_distribution}.
This finding supports the notion that $M_{\text{synth}}$ has a preference for generating common knowledge, while the knowledge that $M_{\text{train}}$ needs to acquire during training often resides in the rare, long-tail data. To further investigate the relationship between long-tail data and training effectiveness, we explored the scaling law of data generated by GraphGen. Similar to the concept of  hard example mining \citep{shrivastava2016training}, we sorted the synthetic data in descending order of $\text{Loss}_C$
and divided it into different proportions for sequential training to emphasize the importance of focusing on the most challenging instances. Surprisingly, we found that even when trained on only a small proportion of data (less than 5\%), the model can still maintain a relatively high proportion of performance, as can be seen in Figure~\ref{exp:scaling}. As the total amount of training data increases, the overall score shows minimal improvement. Therefore, the head of the generated data contributes little new knowledge to the model.

Additionally, we conducted a comparative experiment by training the model using the top 30\% and bottom 30\% of the data sorted according to $\text{Loss}_C$. The results showed that the model trained on the top 30\% data achieved better performance than that trained on the bottom 30\% data. 
In our study, comprehension loss is the discrepancy between a model’s confidence in predicting correct or incorrect statements and the actual ground-truth accuracy. Higher comprehension loss values indicate knowledge blind spots within the model. These high-loss instances often involve long-tail or rare knowledge that the model may struggle with. Training on high-loss data allows models to learn from knowledge underrepresented in earlier training stages. 
Although this type of knowledge is rare, it is critical for knowledge-intensive tasks and can lead to performance improvements. 
This finding demonstrates that data with higher $\text{Loss}_C$ can bring greater performance gains to the Trainee Model, as can be seen in Appendix~\ref{appendix:effect_of_comprehension}.

\subsection{Comprehension Loss Change}


\begin{figure}[t]
\centering
  \includegraphics[width=\linewidth]{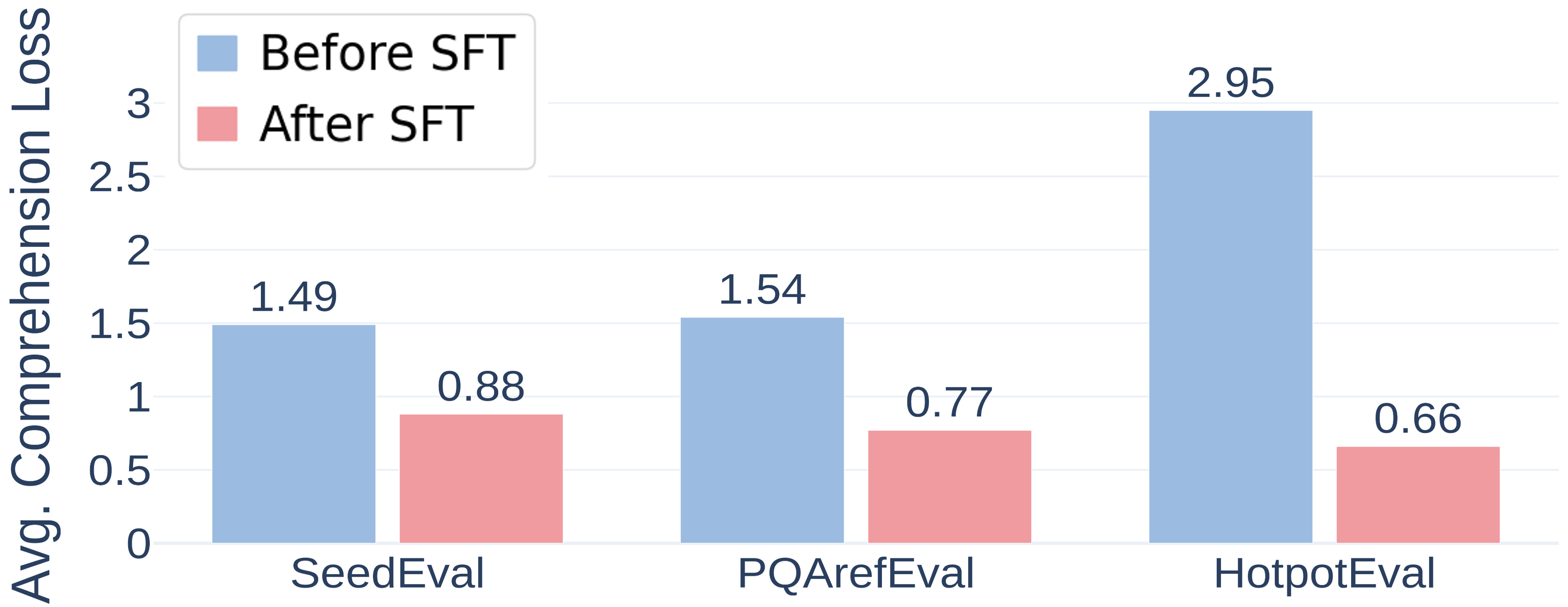}  \caption{\label{exp:loss_decrease}Comprehension loss of the Trainee Model. The reduction in loss after training highlights the effectiveness of data synthesis and the enhanced comprehension ability of the Trainee Model.}
\end{figure}


After the SFT phase, we conducted a comprehension assessment on $M_f$. Although we did not directly use yes/no judgment questions as part of the training data, we observed a significant reduction in $\text{Loss}_C$, as can be seen in Figure~\ref{exp:loss_decrease}. This indicates that GraphGen has enhanced $M_{\text{train}}$'s understanding of the knowledge domain, enabling it to reliably differentiate between correct and incorrect statements. Consequently, the model shows greater accuracy in knowledge-intensive tasks.

\subsection{Ablation Studies}

\paragraph{Selection of Entities and Relationships}
In atomic QA generation experiments, 
we compared the effectiveness of using only entities versus only relationships as the sources for generation. The results indicate that relying solely on relationships outperformed using entities alone and even slightly exceeded the performance of using the entire KG. We attribute this to the fact that relationships more effectively encapsulate the intrinsic properties of knowledge. Additionally, the presence of overlap in knowledge organization within the KG may contribute to a decline in performance. The results can be seen in Appendix~\ref{appendix:ablation_study_on_the_selection_strategy}. \\
\paragraph{Selection of Graph Organization Strategies}
We changed the length strategy of GraphGen by setting $pre\_length$ to 256, 512, 768, and 1024, and conducted evaluations on quality metrics for each case. The results can be seen in  Appendix~\ref{appendix:ablation_study_on_the_selection_strategy}. We found that, although the average length of the generated data increased, the final score tended to stabilize. This indicates that the score is not directly correlated with the data length.
We also evaluated the data on $D_{\text{eval}}$, as can be seen in Appendix~\ref{appendix:ablation_study_on_the_selection_strategy}. 
We found that although a longer $pre\_length$ may enhance the long-text ability of large models, the evaluation results with a $pre\_length$ of 256 were the best. The analysis of the length distribution of the final generated data revealed that a potential explanation lies in the characteristics of the data length distribution. Specifically, with a $pre\_length$ of 256, the distribution displays sharper traits for lengths below 5000. In contrast, an increased $pre\_length$ leads to a distribution with greater extension in length. An excessive amount of lengthy data may significantly prolong the convergence time required for the model. 
We also conducted experiments using the selection strategy as control variables. The analysis indicated that the influence of the strategy on the results was minimal, as demonstrated in Appendix~\ref{appendix:ablation_study_on_the_selection_strategy}. This finding suggests that, as long as a correlation exists, variations in understanding levels within the subgraphs do not significantly impact the final outcomes. However, the underlying patterns merit further exploration in future research.

\section{Conclusion}
In this paper, we propose GraphGen, an effective KG-based approach to synthetic data generation for fine-tuning LLMs on knowledge-intensive tasks in closed-book QA settings. GraphGen is specifically designed to meet the needs of three scenarios: atomic QA, aggregated QA, and multi-hop QA. 
Experiments demonstrate that GraphGen successfully addresses limitations of existing synthetic data generation methods by leveraging KGs to guide the creation of high-quality QA pairs. 
Our approach ensures that the generated data is both relevant and diverse, offering a promising solution to 
effectively addressing the data bottlenecks frequently encountered in supervised fine-tuning of LLMs.

\textbf{Future research} could focus on enhancing the knowledge graph construction by integrating external knowledge sources and dynamic updates to improve data quality and coverage. What's more, exploring adaptive graph organization strategies and subgraph sampling methods could optimize the training data for better model performance.

\clearpage

\section*{Limitations}
Despite the promising results exhibited by GraphGen, several limitations necessitate further investigation and enhancement. One significant concern is the framework's requirement for substantial computational resources when building and processing large-scale KGs. This demand may restrict its applicability in resource-constrained environments or when dealing with extensive datasets. Therefore, optimizing computational efficiency is vital for broader adoption.

While GraphGen has demonstrated strong performance across three representative domains, its adaptability to diverse fields remains to be explored. Current experiments have primarily focused on closed-book QA tasks, leaving the framework's generalization to other areas—such as mathematics, reasoning, and coding—largely unexamined. Furthermore, the integration of synthetic data generated by GraphGen into model training requires meticulous tuning. It is essential to investigate the balance between synthetic and real data, as well as their effects on model convergence and generalization.

In this article, we do not discuss open-book question answering, such as Retrieval-Augmented Generation (RAG). RAG's effectiveness is contingent upon the quality and recall capacity of the retrieval corpus, while failures in retrieval can potentially exacerbate the incidence of hallucinations. The integration of data synthesis with RAG methodologies signifies a promising avenue for future in-depth research.


\section*{Acknowledgments}
This work was supported by Shanghai Artificial Intelligence Laboratory. The authors thank Songyang Zhang from Shanghai Artificial Intelligence Laboratory for academic discussion. The authors also thank Silicon Flow\footnote{\url{https://siliconflow.cn/}} for API support.

\clearpage

\bibliography{custom}

\newpage

\appendix
\onecolumn

\section{Additional System Modules} \label{appendix:system}
\subsection{Entity Enrichment Module with Wikipedia}
We have developed a plug-in module for GraphGen aimed at enriching entity information within the KG through targeted Wikipedia searches.  Entities from $D_{\text{source}}$ may initially contain only rudimentary descriptions. However, by leveraging the extensive and authoritative content of Wikipedia, we can provide a more comprehensive understanding of these entities, including historical context, definitions, and related facts. This enhancement not only enriches the content of the knowledge graph but also serves as a means to verify the accuracy of existing information, identifying potential errors or gaps within the data.

\subsection{Coreference Resolution Module}
We have developed an additional plug-in module for coreference resolution that processes text segments from $D_{\text{source}}$. By designating the first segment as a reference point, we analyze subsequent segments to identify any ambiguous pronouns. Utilizing a large language model (LLM), we generate responses that clarify these references based on the context established by the reference segment. This approach enables us to accurately identify and resolve pronouns and other referring expressions, thereby enhancing the clarity and coherence of the text segments.

\subsection{User Interface}
\begin{figure}[h!]
  \includegraphics[width=\textwidth]{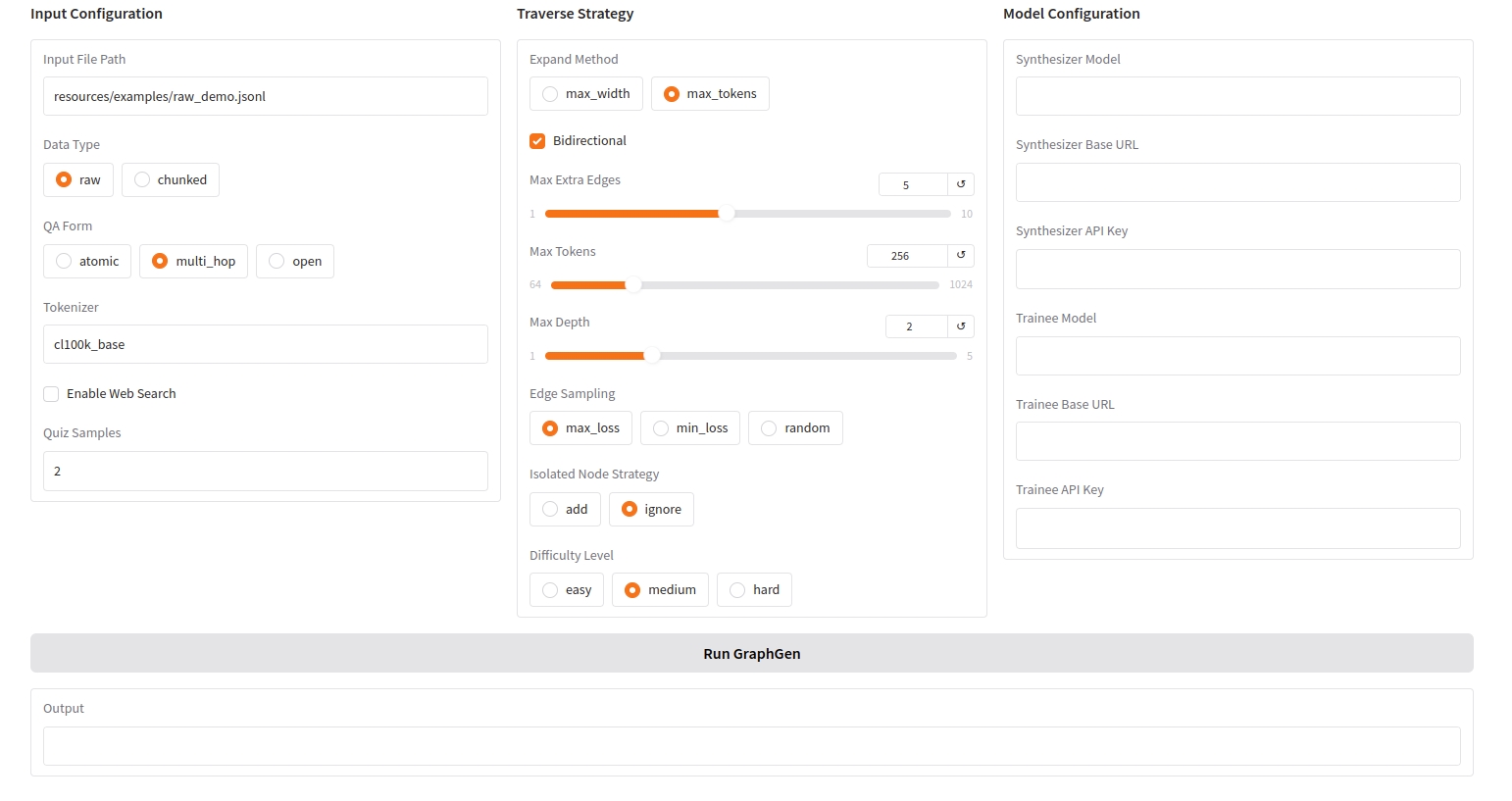}
  \caption{\label{appendix:inerface}User interface of GraphGen. ``Input Configuration'' is utilized to specify the data sources and the target format. ``Traverse Strategy'' determines the method of graph organization. ``Model Configuration'' is employed to set parameters for the LLMs.}
\end{figure}

The user interface of GraphGen is designed to provide users with an intuitive tool for modifying and adjusting various settings. As can be seen in Figure~\ref{appendix:inerface}, the settings include ``Input Configuration'', ``Traverse Strategy'' and ``Model Configuration''. Users can save their current parameter configurations as presets for easy retrieval in the future. This is especially useful for those who frequently adjust settings, enhancing efficiency.

\newpage
\section{GraphGen Details} \label{appendix:graphgen_details}
\subsection{Prompt Templates}

\begin{figure}[!]
  \includegraphics[width=0.95\textwidth]{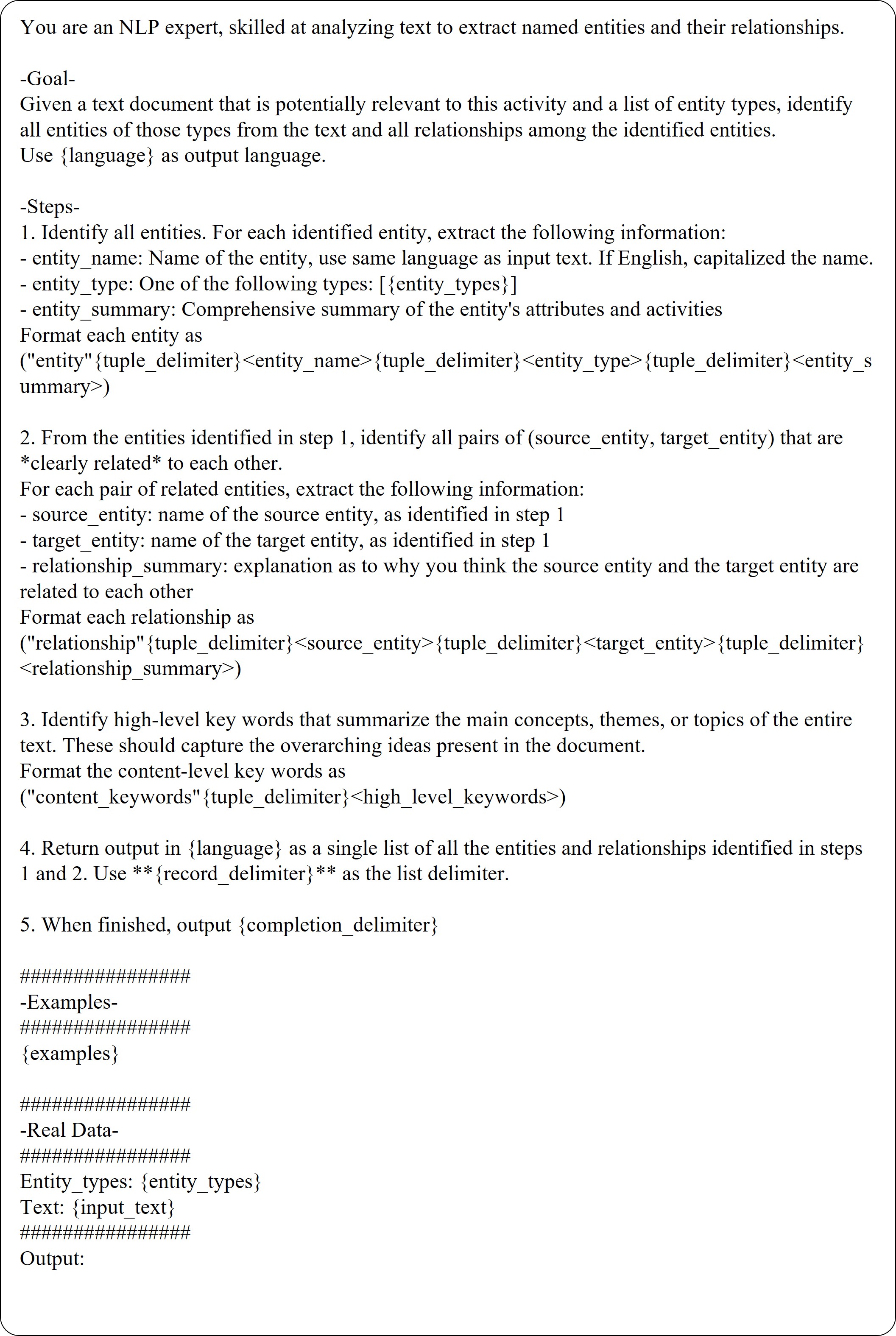}
  \caption{\label{prompt:kg_extraction}Prompt for KG extraction.}
\end{figure}

\begin{figure}[!]
  \includegraphics[width=0.95\textwidth]{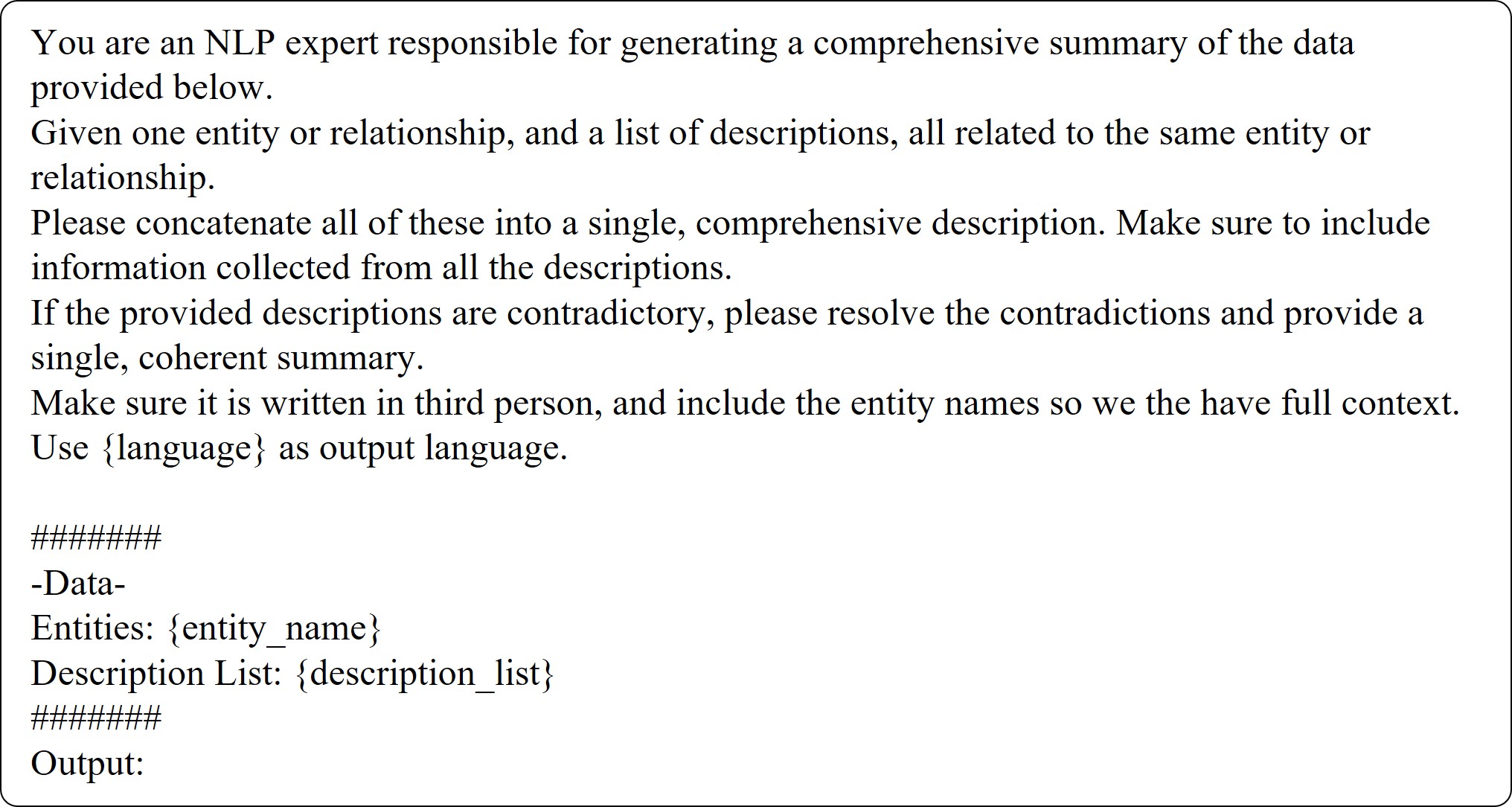}
  \caption{\label{prompt:kg_summarization}Prompt for KG summarization.}
\end{figure}

\begin{figure}[!]
  \includegraphics[width=0.95\textwidth]{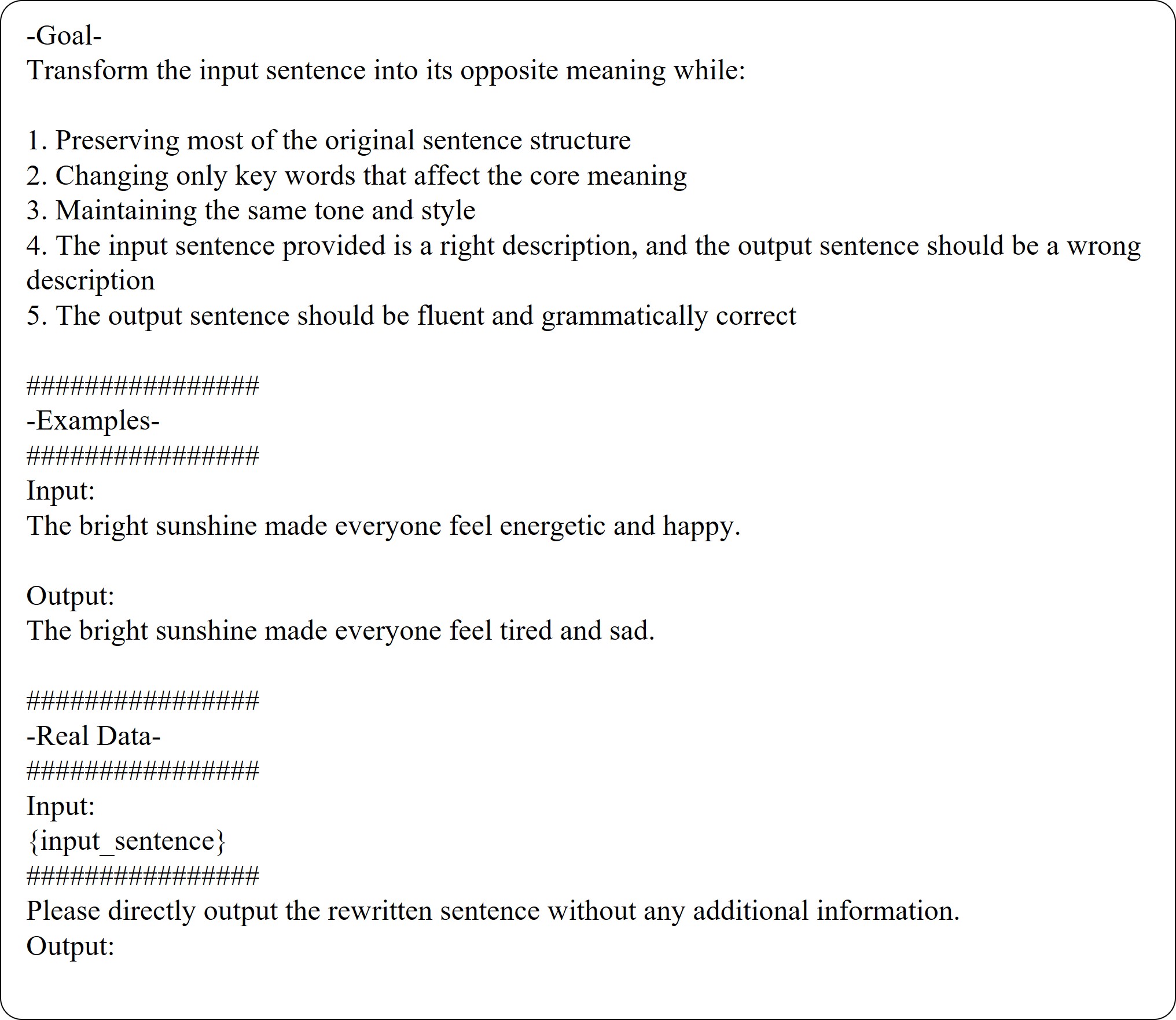}
  \caption{\label{prompt:description_rephrasing(opposite)}Prompt for description rephrasing (opposite meaning).}
\end{figure}

\begin{figure}[!]
  \includegraphics[width=0.95\textwidth]{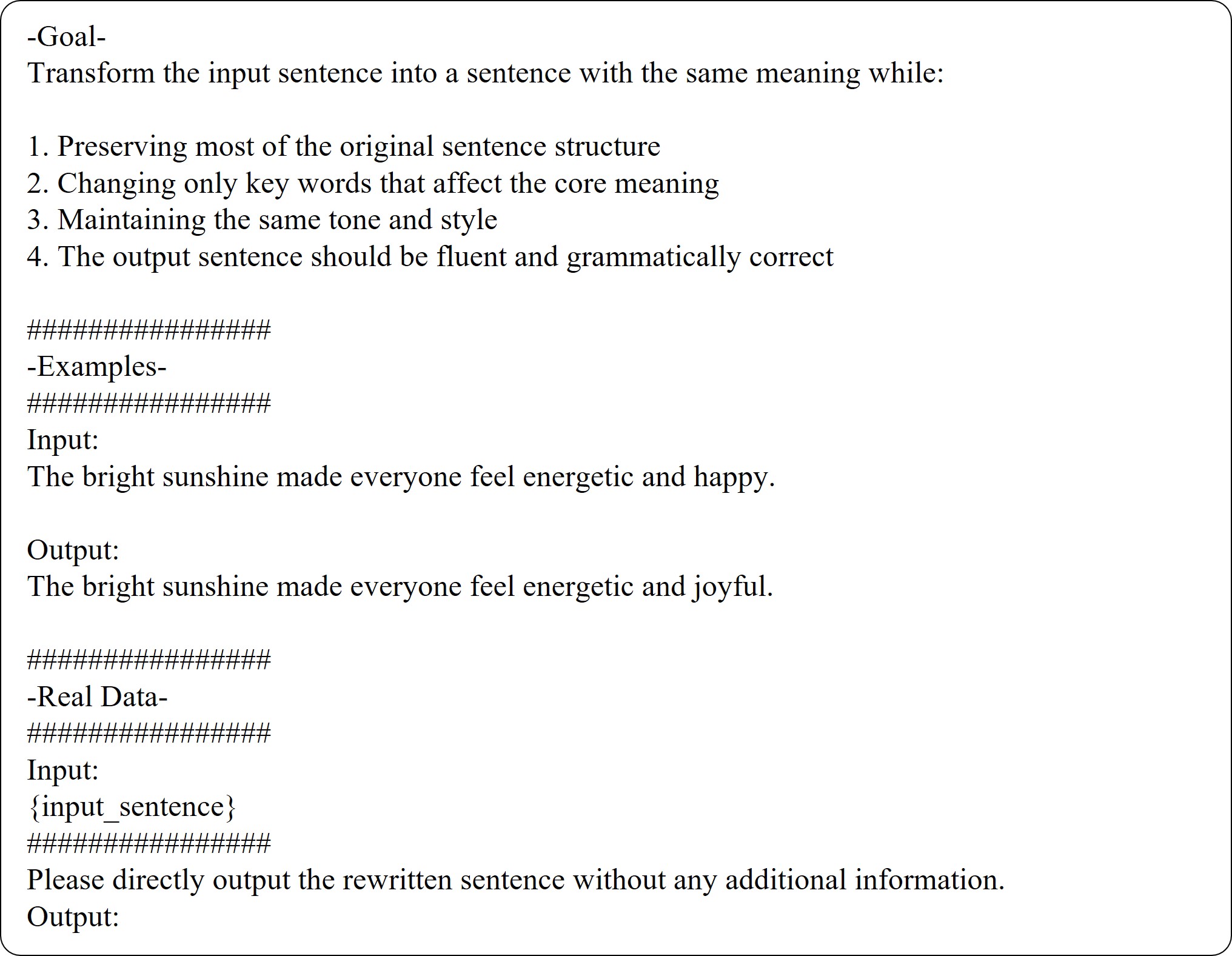}
  \caption{\label{prompt:description_rephrasing(literal)} Prompt for description rephrasing (literal meaning).}
\end{figure}

\begin{figure}[!]
  \includegraphics[width=0.95\textwidth]{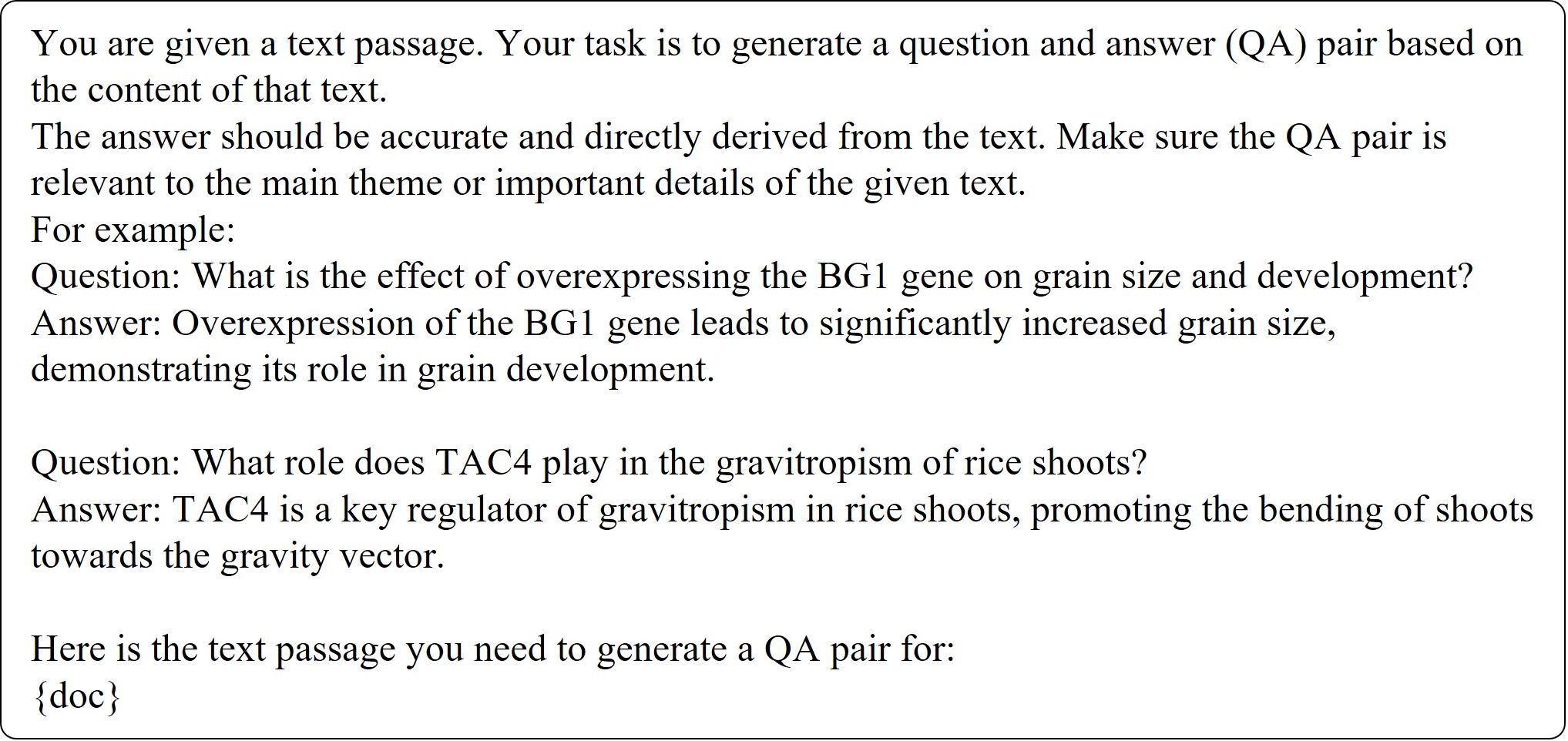}
  \caption{\label{prompt:atomic_question_generation}Prompt for atomic QA generation.}
\end{figure}

\begin{figure}[!]
  \includegraphics[width=0.95\textwidth]{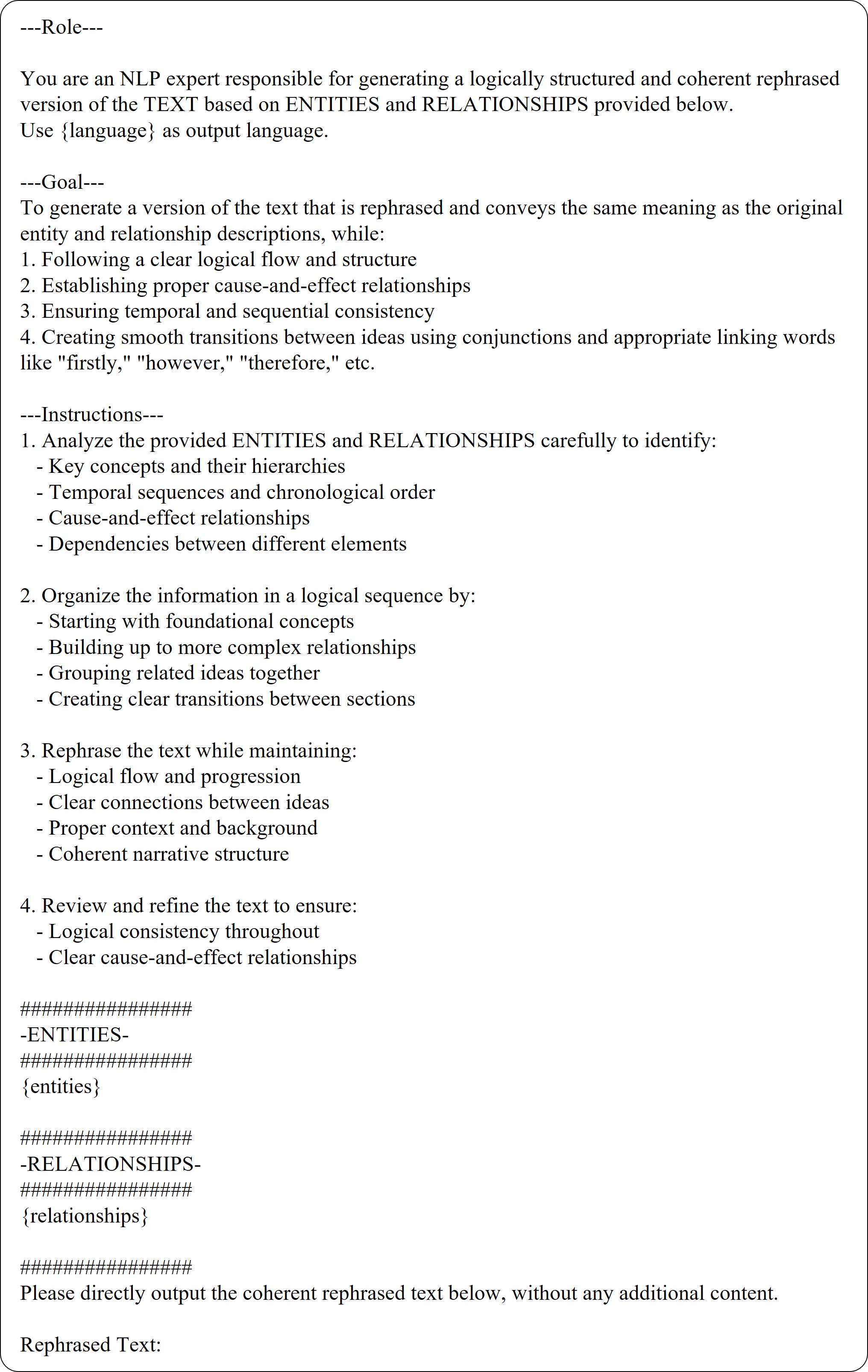}
  \caption{\label{prompt:answer_rephrasing}Prompt for aggregated answer rephrasing.}
\end{figure}

\begin{figure}[!]
  \includegraphics[width=0.95\textwidth]{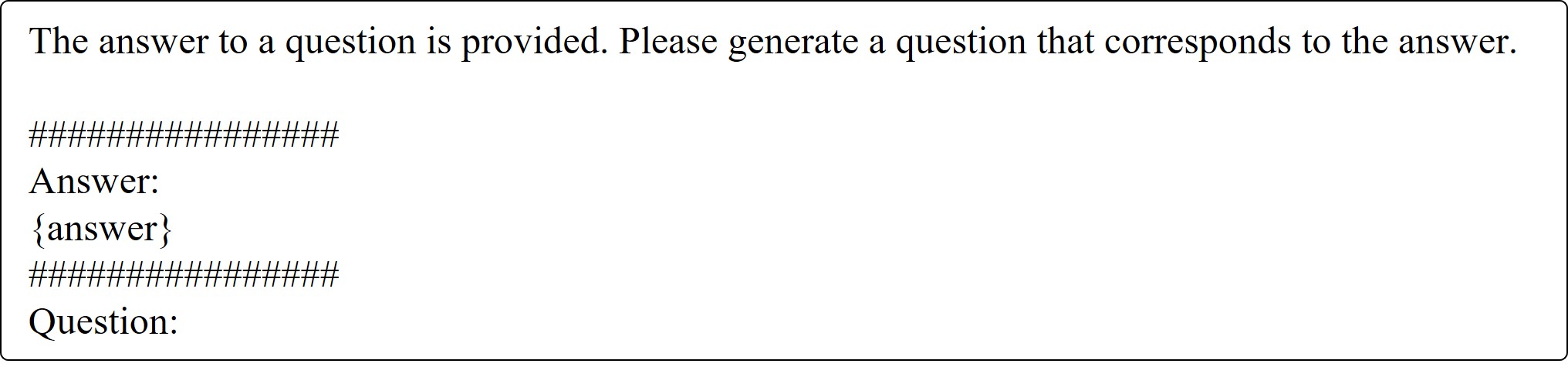}
  \caption{\label{prompt:question_generation}Prompt for question generation (aggregated QA).}
\end{figure}

\begin{figure}[!]
  \includegraphics[width=0.95\textwidth]{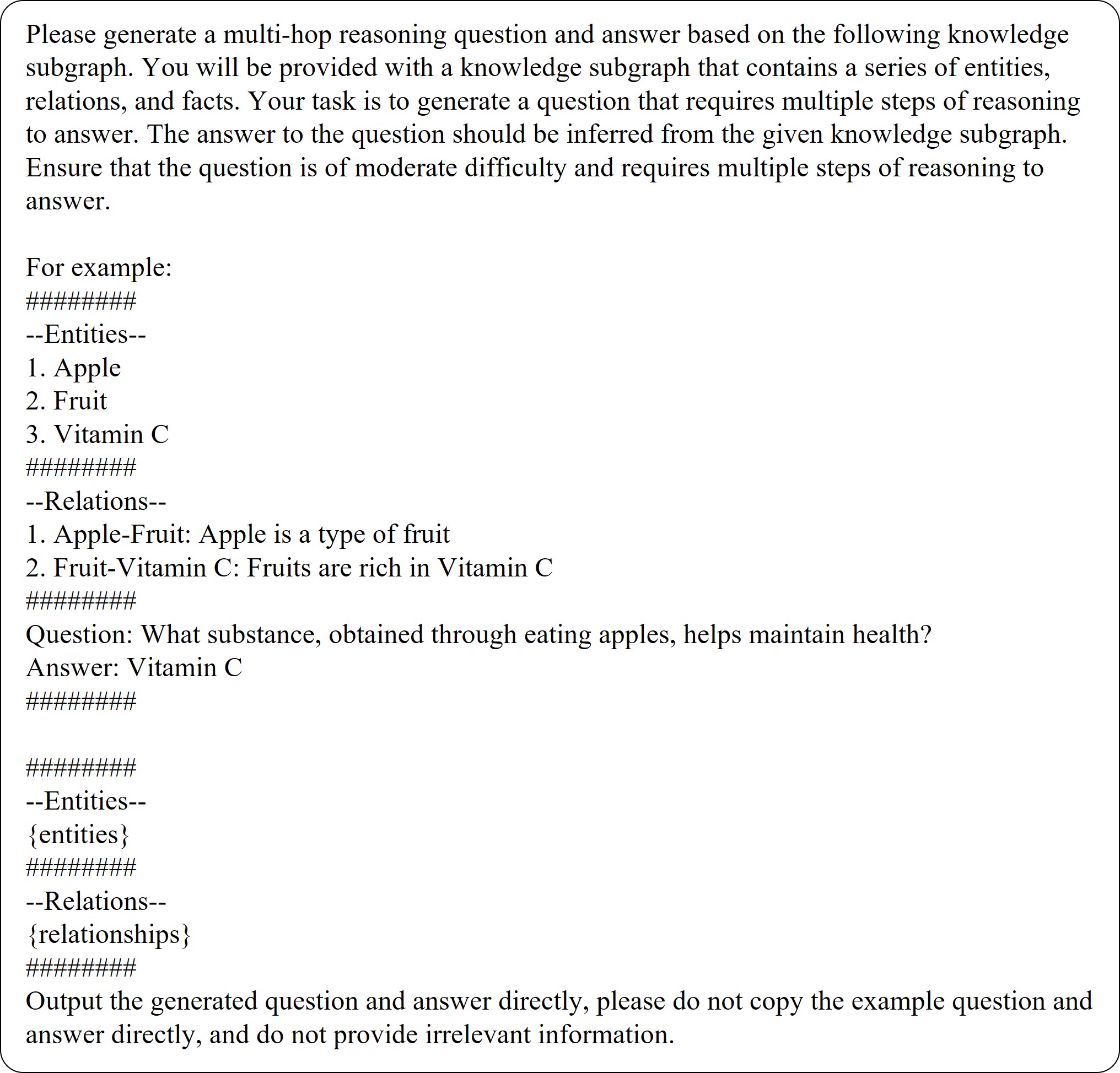}
  \caption{\label{prompt:multi_hop_generation}Prompt for multi-hop QA generation.}
\end{figure}

In the GraphGen framework, we used the following prompt:
\begin{itemize}
    \item Prompt for extracting entities and relationships of the KG (Figure~\ref{prompt:kg_extraction}).
    \item Prompt for summarizing multiple descriptions when the descriptions of an entity or relation come from various sources (Figure~\ref{prompt:kg_summarization}).
    \item Prompt for rephrasing the description into a positive or a negative statement (Figure~\ref{prompt:description_rephrasing(opposite)} and ~\ref{prompt:description_rephrasing(literal)}).
    \item Prompt for atomic QA generation (Figure~\ref{prompt:atomic_question_generation}).
    \item Prompt for aggregated QA generation (Figure~\ref{prompt:answer_rephrasing} and ~\ref{prompt:question_generation}).
    \item Prompt for multi-hop QA generation (Figure~\ref{prompt:multi_hop_generation}).
\end{itemize}

\newpage

\newpage

\newpage

\newpage

\newpage

\newpage
\subsection{Examples}
\begin{figure}[!]
\centering
  \includegraphics[width=0.7\textwidth]{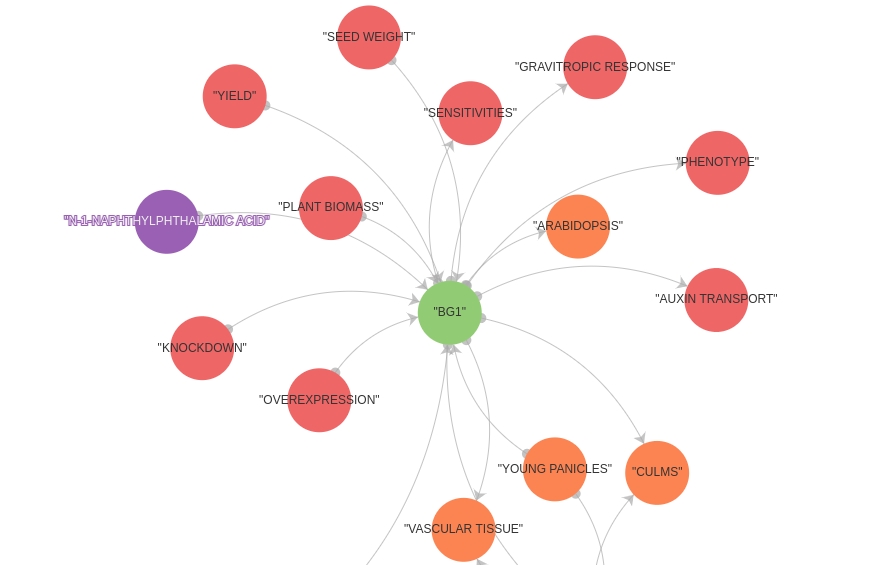}
  \caption{\label{appendix:graph}An example of the extracted KG. Different colors represent different entity types.}
\end{figure}

Here we present some output examples from the GraphGen workflow. Figure~\ref{appendix:graph} shows an example of the extracted KG. In the graph, entities are interconnected based on the relationships obtained from $D_{\text{source}}$, and each entity or relation has its own description. 

\begin{figure}[!]
  \centering
  \includegraphics[width=0.9\textwidth]{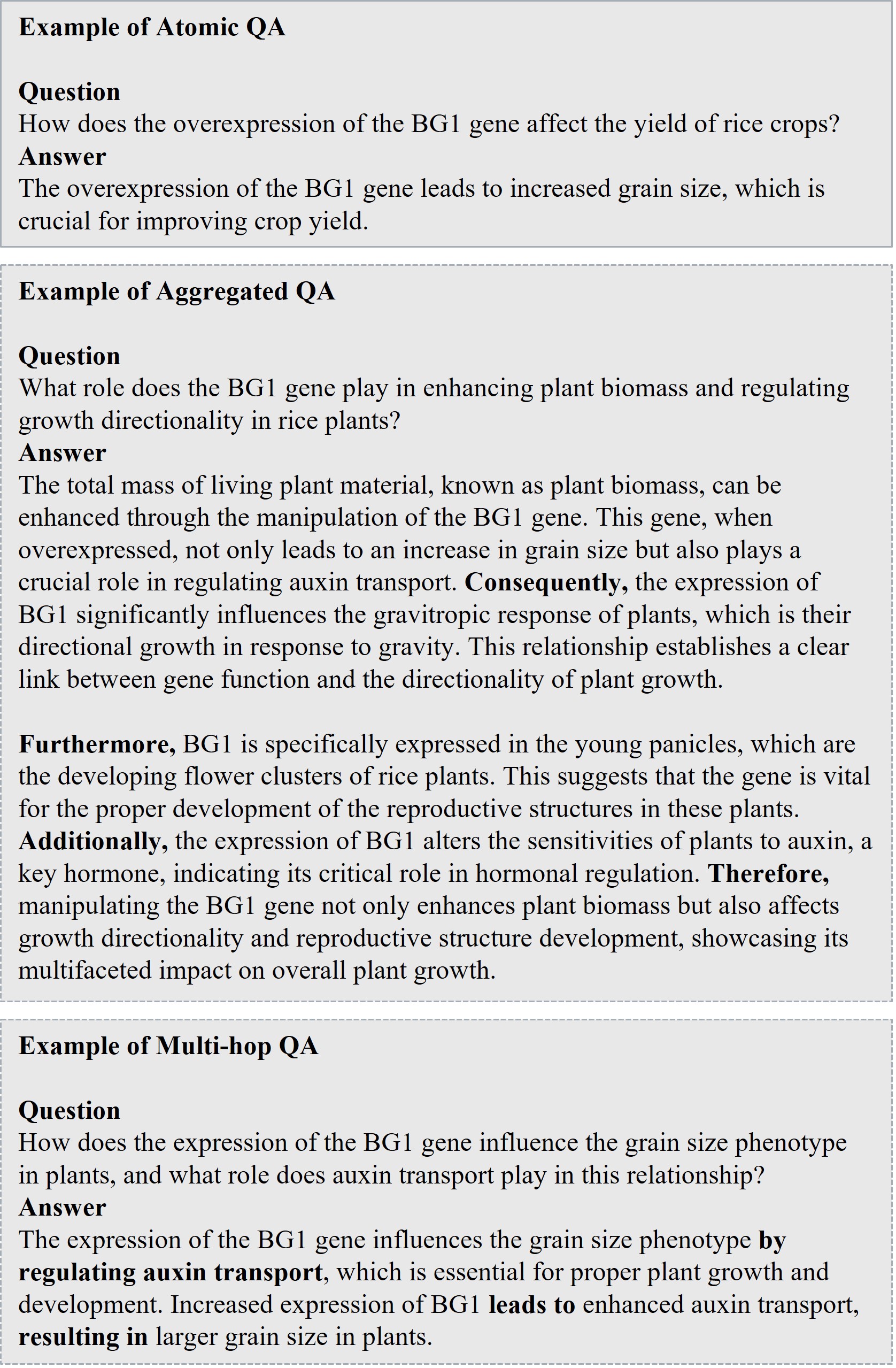}
  \caption{\label{appendix:output_example}Examples of the GraphGen data.  The words indicating contrasts or clear logical relationships between knowledge points is highlighted.}
\end{figure}

Figure~\ref{appendix:output_example} illustrates the three styles of data generated by GraphGen. We can clearly observe that atomic QA focuses on simple, single knowledge points, while aggregated QA generates a coherent and logical long answer within complex subgraphs, thereby producing more complex and comprehensive responses. Multi-hop QA, on the other hand, emphasizes reasoning and connecting multiple knowledge points. These methods each demonstrate different levels of knowledge extraction and semantic understanding capabilities.

\newpage

\newpage
\subsection{Implementation Details} \label{appendix:implementation_details}
Here we present the complete configuration of GraphGen's organization strategy, as can be seen in Table~\ref{tab:traversal-strategy}.
\begin{table}[h]
    \centering
    \small
    \begin{tabular}{l p{3cm} p{3cm} p{5cm}}
        \toprule
        \textbf{Parameter} & \textbf{Default Value} & \textbf{Description} & \textbf{Options} \\
        \midrule
        qa\_form & atomic & Type of QA form desired. &
        \begin{minipage}[t]{\linewidth}
        \begin{itemize}[itemsep=0pt, parsep=0pt]
        \item atomic: single-step
        \item multi\_hop: multi-step
        \item aggregated: open-ended
        \end{itemize}
        \end{minipage} \\
        expand\_method & max\_tokens & Method for controlling graph expansion. & 
        \begin{minipage}[t]{\linewidth}
        \begin{itemize}[itemsep=0pt, parsep=0pt]
        \item max\_width: limit by number of edges in the subgraph
        \item max\_tokens: limit by token length of entities and relations in the subgraph
        \end{itemize}
        \end{minipage}
        \\
        bidirectional & True & Expanding the graph in both directions (True) or one direction (False). \\
        max\_extra\_edges & 5 & Maximum number of edges to expand. \\
        max\_tokens & 256 & Maximum number of tokens. \\
        max\_depth & 2 & Maximum depth for traversal in each direction. \\
        edge\_sampling & max\_loss & Strategy for edge selection at the same layer. &
        \begin{minipage}[t]{\linewidth}
        \begin{itemize}[itemsep=0pt, parsep=0pt]
        \item max\_loss: prioritize highest loss edges
        \item min\_loss: prioritize lowest loss edges
        \item random: random selection
        \end{itemize}
        \end{minipage} \\
        isolated\_node\_strategy & add & Handling strategy for isolated nodes. &
        \begin{minipage}[t]{\linewidth}
        \begin{itemize}[itemsep=0pt, parsep=0pt]
        \item add: include isolated nodes
        \item ignore: exclude isolated nodes
        \end{itemize}
        \end{minipage}
        \\
        \bottomrule
    \end{tabular}
    \caption{Configuration of the graph organization strategy.}
    \label{tab:traversal-strategy}
\end{table}

\newpage
\section{Additional Setups} \label{appendix:additional_setups}


In this section, we present the detailed configurations for generating, training and evaluation settings. Table~\ref{tab:training_config} provides the hyperparameters employed during training, while Table~\ref{tab:evaluation_config} outlines the parameters used in our evaluation pipeline. The time required for processing varies with the size of the dataset and changes in the graph organization strategy. On average, generating a batch of approximately 50,000 data entries takes about 2 hours for Qwen2.5-72B-Instruct\footnote{\url{https://huggingface.co/Qwen/Qwen2.5-72B-Instruct}}, while SFT on Qwen2.5-7B-Instruct\footnote{\url{https://huggingface.co/Qwen/Qwen2.5-7B-Instruct}} requires around 1 hour and evaluation takes about 10 minutes, utilizing 8 NVIDIA A100 40GB GPUs.

When generating data, for $M_{\text{synth}}$, we set the following parameters: $\text{topk} = 50$, $\text{topp} = 0.95$, $\text{repetition\_penalty} = 1.05$, $\text{max\_tokens} = 10240$, and $\text{temperature} = 0$. It is noteworthy that when rephrasing descriptions, the temperature is adjusted to 1 for diverse expressions. When judging statements, for $M_train$, we need to obtain the softmax probabilities of the output tokens. Therefore, we set the parameters as follows: $\text{logprobs} = \text{True}$, $\text{top\_logprobs} = 5$, and $\text{max\_tokens} = 1$.

For evaluation, we leverage the OpenCompass framework~\citep{2023opencompass} as a standardized evaluation toolkit. The evaluation process is controlled through key hyperparameters that dictate model behavior, including sequence length constraints, batch processing, and sampling configurations. The detailed parameter settings are presented in Table~\ref{tab:evaluation_config}.

\begin{table}[h!]
    \centering
    \begin{tabular}{l p{2.5cm} p{8cm}}
        \toprule
        \textbf{Parameter} & \textbf{Value} & \textbf{Description} \\
        \midrule
        \cmidrule(r){1-3}
        Maximum Sequence Length & 7168 & The maximum length of input tokens the model can process in a single instance. \\
        Maximum Output Length & 2048 & The maximum number of newly generated tokens allowed in model output. \\
        Batch Size & 80 & The maximum number of prompts that LMDeploy receives in the `generate' function. \\
        Temperature (Gen.) & 0 & Controls randomness in sampling; lower values lead to more deterministic outputs (greedy search). A value of 0 enforces fully deterministic generation. \\
        \bottomrule
    \end{tabular}
    \caption{Evaluation configuration parameters.}
    \label{tab:evaluation_config}
\end{table}

\newpage

In the training phase, we employ a transformer-based architecture with the AdamW optimizer. The learning rate is linearly scheduled with a warm-up phase, and gradient clipping is applied to stabilize training. These configurations ensure effective optimization and robust model convergence.

\begin{table}[h!]
    \centering
    \begin{tabular}{l p{2.5cm} p{8cm}}
        \toprule
        \textbf{Parameter} & \textbf{Value} & \textbf{Description} \\
        \midrule
        \cmidrule(r){1-3}
        Maximum Length & 2048 & Maximum sequence length for model inputs. \\
        Learning Rate & 2e-5 & Step size for weight updates, controlling optimization speed. \\
        Weight Decay & 0.1 & Regularization term to mitigate overfitting by penalizing large weights. \\
        Gradient Clipping & 1 & Caps gradient norm to stabilize training and prevent exploding gradients. \\
        Batch Size & 64 (16$\times$4) & Total number of samples processed per optimization step. \\
        Optimizer & AdamW & Variant of Adam with decoupled weight decay for improved generalization. \\
        Betas ($\beta_1, \beta_2$) & (0.9, 0.999) & Exponential decay rates for first and second moment estimates in Adam. \\
        Warmup Ratio & 0.03 & Fraction of total training steps used for gradual learning rate ramp-up. \\
        Number of Epochs & 2 & Total number of complete passes over the training dataset. \\
        \bottomrule
    \end{tabular}
    \caption{Training configuration parameters.}
    \label{tab:training_config}
\end{table}

\newpage
\section{Baseline Details} \label{appendix:baseline_details}

Among the baseline methods, WRAP, Genie, LongForm, and SELF-QA are generation methods based on prompt engineering, while EntiGraph is based on KG. Specifically, LongForm utilizes the text segments from $D_{\text{source}}$ directly as answers in QA pairs, subsequently generating corresponding questions based on these answers. Genie feeds raw text into a LLM to produce a QA pair. WRAP also extracts QA pairs from raw text but varies in the number of QA pairs generated for each text segment. SELF-QA involves two critical steps: first, it generates ten questions based on the original text, and then it answers these questions contextually, yielding a total of ten QA pairs. EntiGraph begins by extracting entities from the text, then combines these entities in pairs or triplets to create QA pairs, informed by the analysis of the original text. To optimize performance and prevent the generation of excessive, redundant information that could waste computational resources, we implemented a limit on the number of entities selected by EntiGraph during its execution.

\newpage
\section{Dataset Details} \label{appendix:dataset_details}
\textit{SeedEval} is adapted from \textit{SeedBench} \footnote{\url{https://anonymous.4open.science/r/SeedBench}}, a benchmark with 11 tasks related to seed knowledge. For this study, we selected Task QA–4 (covering one-shot and zero-shot scenarios) related to textual knowledge question answering.
\textit{PQArefEval} is derived from \textit{PQAref}, from which we extracted 5,818 instances for our analysis.
\textit{HotpotQA} is a dataset for diverse, explainable multi-hop question answering, where questions require integrating information from multiple sources. We used the test set of \textit{HotpotQA} as the new evaluation dataset, \textit{HotpotEval}. 
Each dataset comprises two components: the QA test set ($D_{\text{eval}}$) and the corresponding source texts ($D_{\text{source}}$). The Corpus for \textit{SeedEval} is provided by anonymous agricultural experts, while \textit{PQArefEval} and \textit{HotpotEval} are constructed from the original references of \textit{PQAref} and \textit{HotpotQA}, respectively. 
Table~\ref{tab:dataset_examples} presents examples from the datasets.

\begin{table*}[h!]
  \centering
  \begin{tabular}{l p{6cm} p{5cm}}
    \toprule
    \textbf{Dataset} & \multicolumn{2}{l}{\textbf{Examples}} \\
    \cmidrule(r){2-3}
    & Corpus & Test\\
    \midrule
    \textit{SeedEval} & \multirow{2}{=}{Grain size is one of the key factors determining grain yield. However, it remains largely unknown how grain size is regulated by developmental signals. Here, we report the identification and characterization ...} & \textbf{Question:} What were the findings regarding the effect of elevated DEP1 accumulation on grain length?\\
    & & \textbf{Answer:} Elevated DEP1 accumulation increased the grain length by 6.85–9.58\% with a normal plant stature.\\
    \midrule
    \textit{PQArefEval} & \multirow{2}{=}{Clinical analysis of the acromial height-measuring device combined with new-type clavicular hook plate and standard clavicular hook plate in the treatment of Neer type II distal clavicle fractures. Background: Distal clavicular fracture is a shoulder joint injury that is common in clinical settings and is generally surgically treated using the clavicular hook plate technique with a confirmed curative effect ... } &
    \textbf{Question:} Is Bridging Necessary? \\
    & & \textbf{Answer:} The necessity of bridging depends on the context in which the term is used. In the context of myocardial bridging, it is not a procedure that is done but rather a congenital condition where a segment of a coronary artery takes an intramyocardial course, which can be asymptomatic or lead to complications such as ...
    
    \\
    \midrule
    \textit{HotpotEval} & \multirow{2}{=}{Ed Wood is a 1994 American biographical period comedy-drama film directed and produced by Tim Burton, and starring Johnny Depp as cult filmmaker Ed Wood ...} & \textbf{Question:} The Vermont Catamounts men's soccer team currently competes in a conference that was formerly known as what from 1988 to 1996? \\
    & & \textbf{Answer:} the North Atlantic Conference \\
    \bottomrule
  \end{tabular}
  \caption{\label{tab:dataset_examples}Dataset examples.}
\end{table*}

\newpage
\section{Additional Experimental Results} 
\label{appendix:additional_experimental_results}
\subsection{Main Results on Additional Models}
\begin{figure*}[h!]
  \centering
  \includegraphics[width=0.9\linewidth]{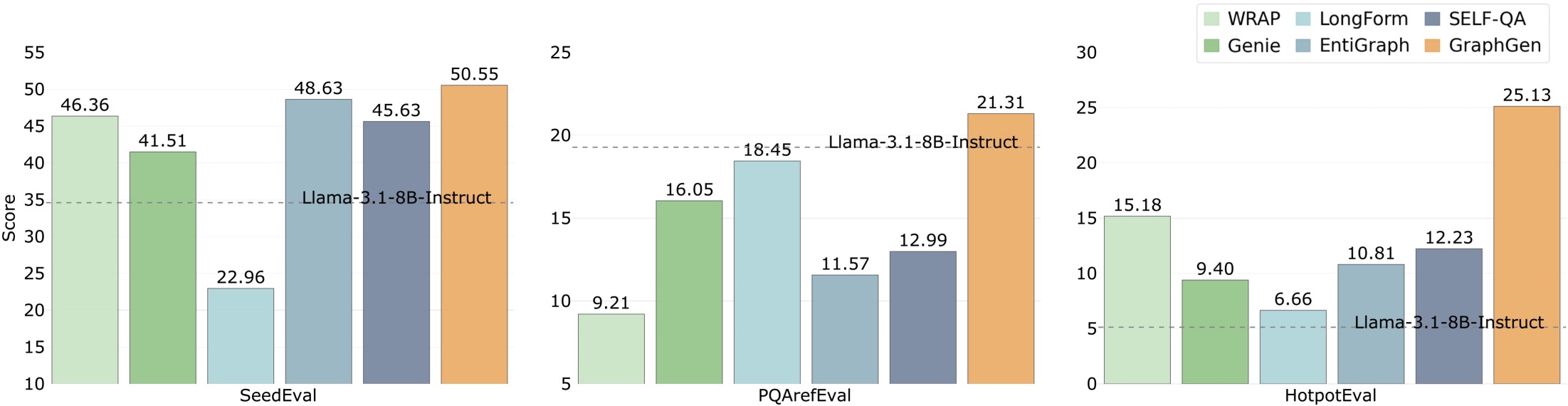}
  \includegraphics[width=0.9\linewidth]{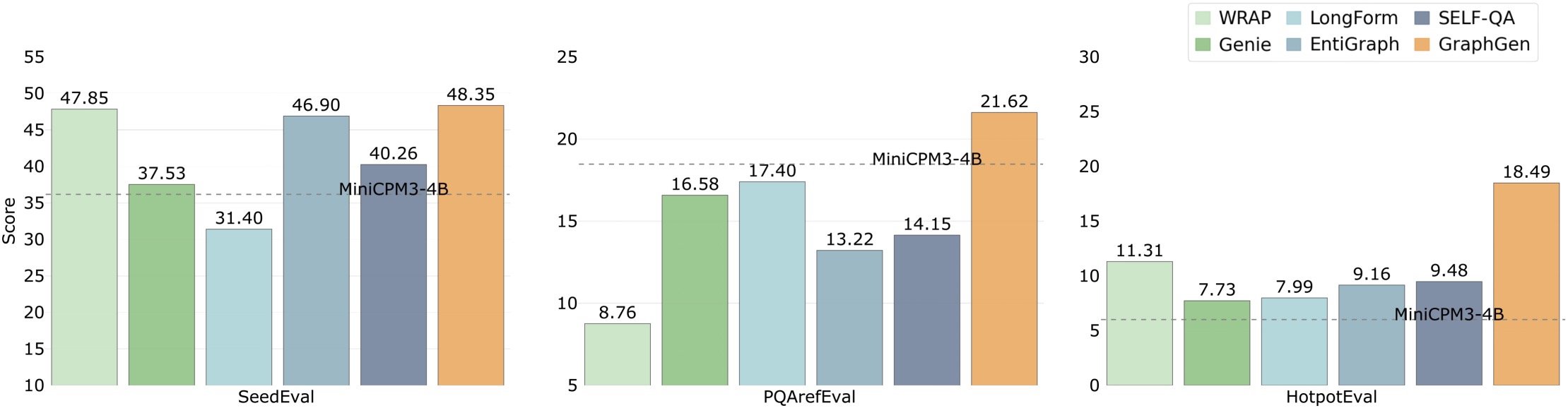}
  \caption{\label{exp:add_eval}Performance comparison on knowledge-intensive evaluation datasets. The models are fine-tuned using data generated by different methods. The top figure shows results on LLaMA-3.1-8B-Instruct, while the bottom figure shows results on MiniCPM3-4B. While baseline methods show varying performance, GraphGen consistently achieves superior results across all three datasets.}
\end{figure*}

To further validate the effectiveness of our proposed method, we conducted additional fine-tuning experiments using two models: Meta-Llama-3.1-8B-Instruct and MiniCPM3-4B, as trainee models. The results, illustrated in Figure~\ref{exp:add_eval}, are consistent with our primary findings. Specifically, our method continues to deliver stable and significant performance improvements across three knowledge-intensive datasets. These additional experiments reinforce the robustness and generalizability of our approach across different LLM architectures and parameter scales.
We have chosen ROUGE-F as the evaluation metric because the evaluation datasets used in this paper consists of question-and-answer problems. ROUGE-F is an evaluation metric used to measure the overlap between the generated text and the reference text, particularly in terms of word-level overlap. It is calculated as the F1 score, which is the harmonic mean of precision and recall. The formula for ROUGE-F is as follows:
$$F1 = 2 \times \frac{Precision \times Recall}{Precision + Recall}$$
\noindent where precision is the proportion of correct words in the generated text relative to the total number of words in the generated text and recall is the proportion of words in the reference text that are correctly predicted relative to the total number of words in the reference text.
We selected ROUGE-F because it provides a comprehensive measure of both precision and recall, making it suitable for evaluating the quality of generated answers in question-and-answer datasets. 

\label{appendix:add_main_experiments}

\newpage
\subsection{Effect of Comprehension Loss on Model Performance} \label{appendix:effect_of_comprehension}
\begin{figure}[h!]
  \centering
  \includegraphics[width=0.6\textwidth]{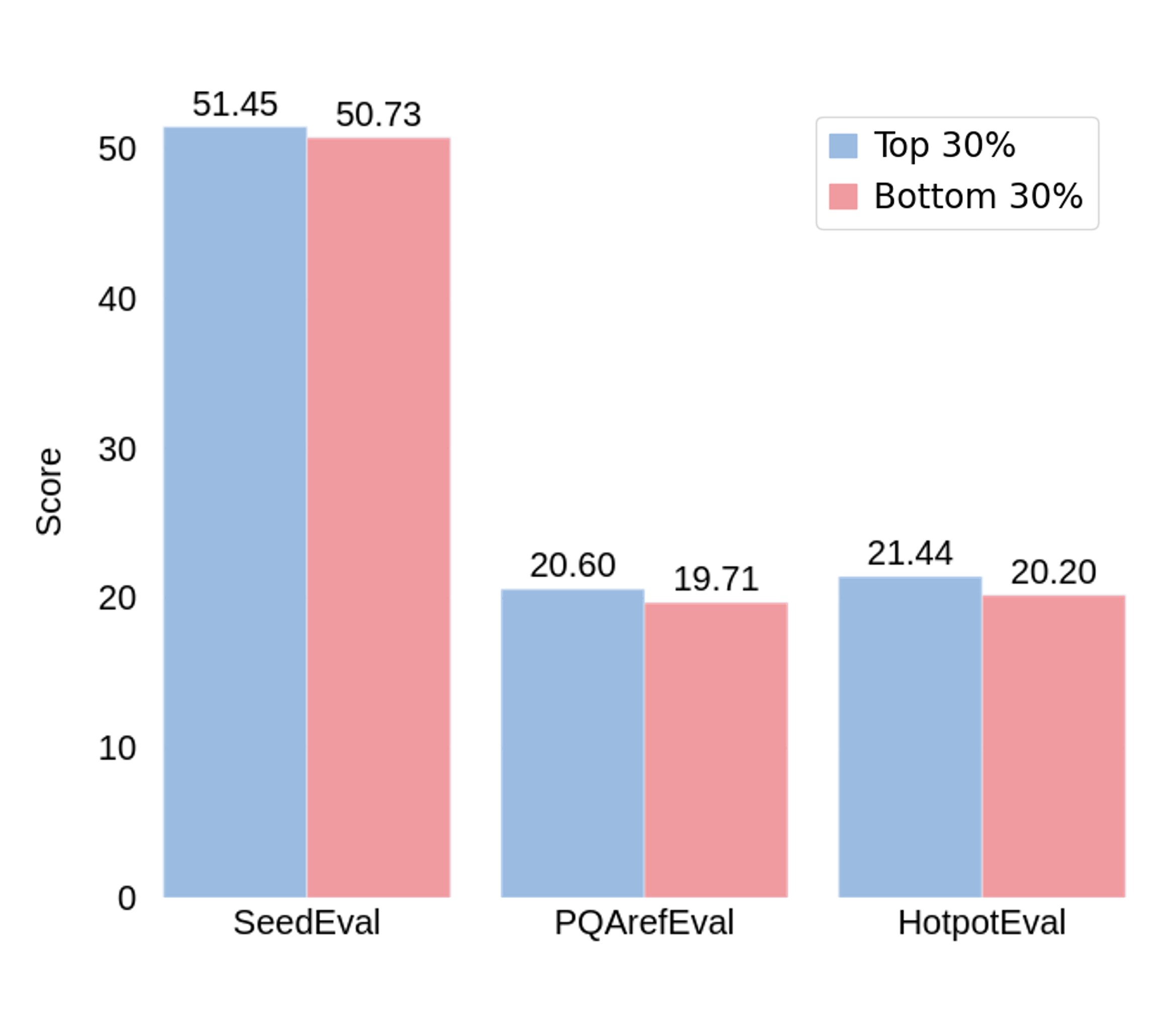}
  \caption{\label{exp:contrast}Comparison of model performance trained on top 30\% and bottom 30\% data sorted by comprehension loss. This figure demonstrates that the model trained on data with higher loss (top 30\%) outperforms the one trained on data with lower loss (bottom 30\%), highlighting the positive impact of high-loss data on model performance enhancement.}
\end{figure}

Figure~\ref{exp:contrast} compares model performance when trained on the top 30\% and bottom 30\% of data sorted by comprehension loss. The results indicate that models trained on higher-loss data achieve better performance, suggesting that such data contributes positively to model optimization and generalization.

\newpage
\subsection{Quality Evaluation Metric Details} \label{appendix:metric_evaluation_details}

Table~\ref{tab:indicator} presents the metrics used to evaluate the intrinsic quality of the text. To normalize the metric scores to a range of 0 to 100, we employ min-max normalization as specified in Formula~\ref{eq:min_max}. In this analysis, the minimum and maximum values for MTLD are established at 0 and 200, respectively. The three metrics included in the Uni-Score are scaled from 0 to 1. For the Reward Score, the range for \textit{Ind} is 0–5, while for \textit{Deb}, it spans from 0 to 3. We use Formula~\ref{eq:average_metric} to compute the final average score $S_{Avg}$, which provides a comprehensive assessment of data quality.

\begin{equation}
  \label{eq:min_max}
  S(x) = \frac{x-x_{min}}{x_{max}-x_{min}} \times 100
\end{equation}

\begin{equation}
  \label{eq:average_metric}
  \begin{aligned}
  S_{Uni} &= \frac{S_{Nat} + S_{Coh} +S_{Und}}{3} \\
  S_{Rew} &= \frac{S_{Ind} +S_{Deb}}{2} \\
  S_{Avg} &= \frac{S_{MTLD} +S_{Uni} + S_{Rew}}{3}
  \end{aligned}
\end{equation}

\begin{table*}[h!]
  \centering
  \begin{threeparttable}[!]
  \resizebox{\textwidth}{!}{
  \begin{tabular}{l l p{11.5cm}}
    \toprule
    \textbf{Metric} & \textbf{Notation} & \textbf{Explanation} \\
    \midrule
     MTLD & $MTLD$ &  The metric for assessing lexical diversity in texts \citep{mccarthy2010mtld}.\\
    Unieval Score & $Uni$ & Naturalness, coherence and understandability score provided by the UniEval dialogue model \citep{zhong2022towards}.\\
     Reward Score & $Rew$ &  The reward model inference score of QA pairs. Reward models in this paper include \textit{BAAI/IndustryCorpus2\_DataRater}\tnote{1} \ and \textit{OpenAssistant/reward-model-deberta-v3-large-v2}\tnote{2}.   \\
    \bottomrule
  \end{tabular}
  } 
   \begin{tablenotes}
     \item[1] \url{https://huggingface.co/BAAI/IndustryCorpus2_DataRater}
     \item[2] \url{https://huggingface.co/OpenAssistant/reward-model-deberta-v3-large-v2}
   \end{tablenotes}
  \end{threeparttable}
    \caption{\label{tab:indicator}Key metrics for evaluating the quality of generated text. The table provides a summary of the notation and explanations for each metric.}
\end{table*}

\newpage
\subsection{Ablation Study on the Selection Strategy} \label{appendix:ablation_study_on_the_selection_strategy}
Tables~\ref{tab:ablation_pqaref} and~\ref{tab:ablation_hotpot} present the results of an ablation study evaluating different edge selection strategies for GraphGen on the \textit{PQArefEval} and \textit{HotpotEval} datasets. The study compares the following strategies: max\_loss, min\_loss, and random. Performance is measured using the ROUGE-F metric.

\begin{table}[h!]
\centering
\begin{tabular}{l c c}
\toprule
\textbf{Method} & \textbf{Selection Strategy} & \textbf{Score} \\
\midrule
\multirow{3}{*}{GraphGen (256)}  
    & max\_loss & 21.14 \\
    & min\_loss & 21.22 \\
    & random & 21.06 \\
\midrule
\multirow{3}{*}{GraphGen (512)}  
    & max\_loss & 20.72 \\
    & min\_loss & 20.92 \\
    & random & 20.93 \\
\midrule
\multirow{3}{*}{GraphGen (768)}  
    & max\_loss & 20.46 \\
    & min\_loss  & 20.47 \\
    & random & 20.64 \\
\midrule
\multirow{3}{*}{GraphGen (1024)}  
    & max\_loss & 20.50 \\
    & min\_loss & 20.56 \\
    & random & 20.40 \\
\bottomrule
\end{tabular}
\caption{Ablation Study on Different Selection Strategies of GraphGen on the \textit{PQAref} Dataset. The model is evaluated under different sequence length settings (pre\_length) and three distinct generation strategies: random, min\_loss, and max\_loss.}
\label{tab:ablation_pqaref}
\end{table}

\begin{table}[h!]
\centering
\begin{tabular}{l c c}
\hline
\textbf{Method} & \textbf{Selction Strategy} & \textbf{Score} \\
\hline
\multirow{3}{*}{GraphGen} & max\_loss  & 23.55 \\
                          & min\_loss  & 23.79 \\
                          & random   & 23.49 \\
\hline
\end{tabular}
\caption{Performance of different generation strategies on the Hotpot dataset. The performance is measured using ROUGE-F.}
\label{tab:ablation_hotpot}
\end{table}

\begin{figure}[htbp!]
    \centering
    \includegraphics[width=0.5\textwidth]{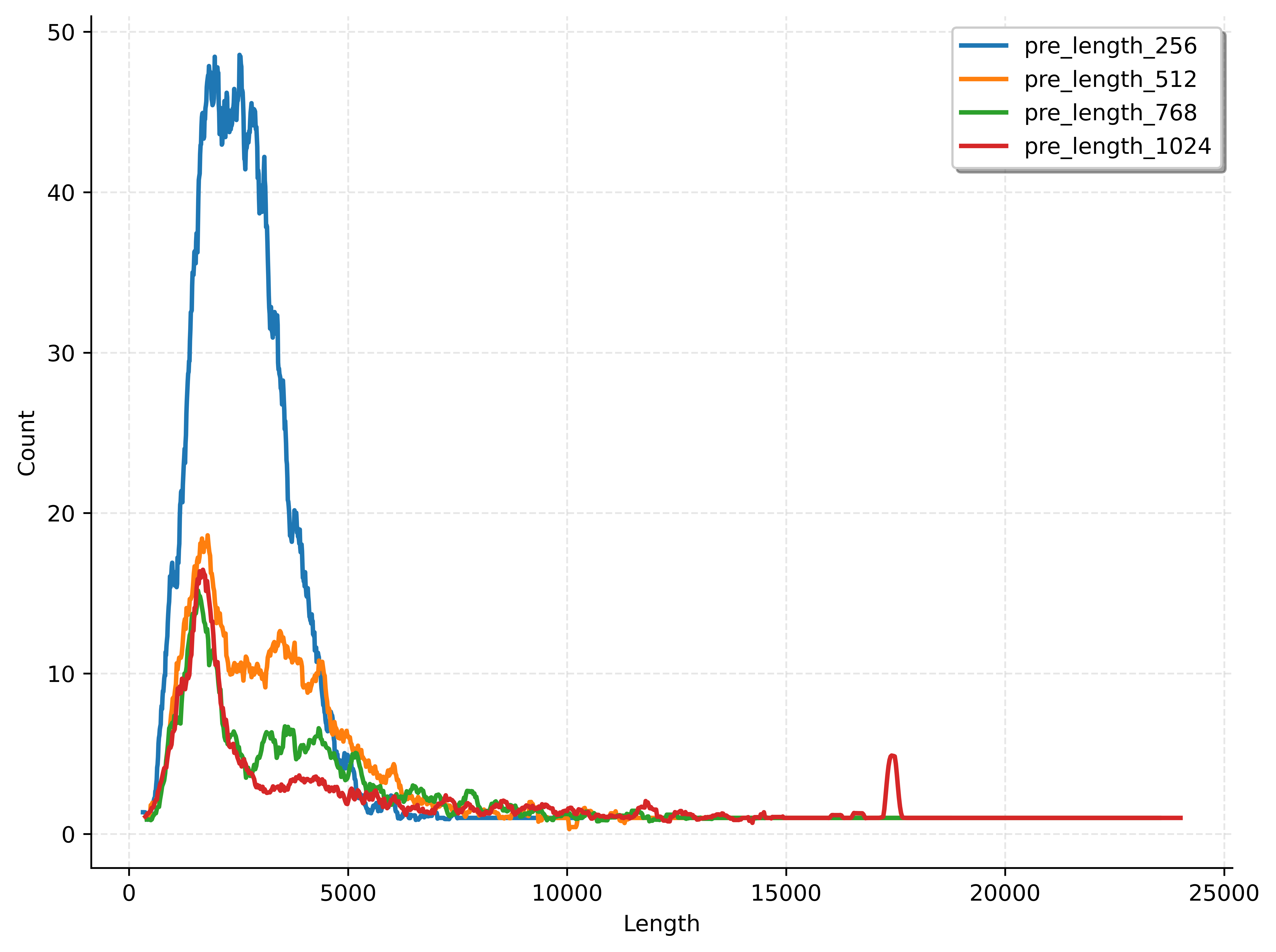}
    \caption{Character length distribution under the maximum loss sampling strategy.}
    \label{fig:max_loss}
\end{figure}

\begin{figure}[htbp!]
    \centering
    \includegraphics[width=0.5\textwidth]{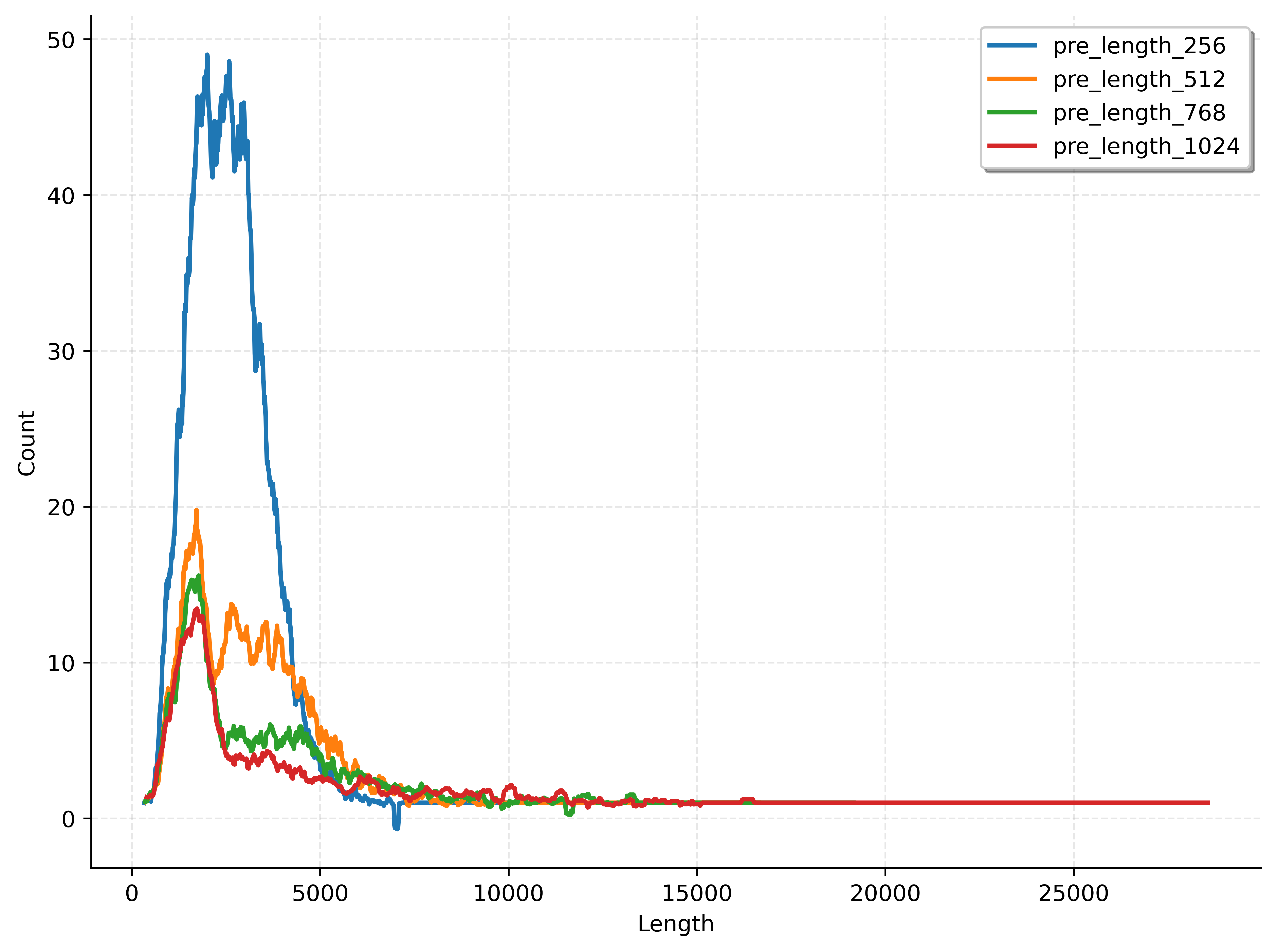}
    \caption{Character length distribution under the minimum loss sampling strategy.}
    \label{fig:min_loss}
\end{figure}

\begin{figure}[H]
    \centering
    \includegraphics[width=0.5\textwidth]{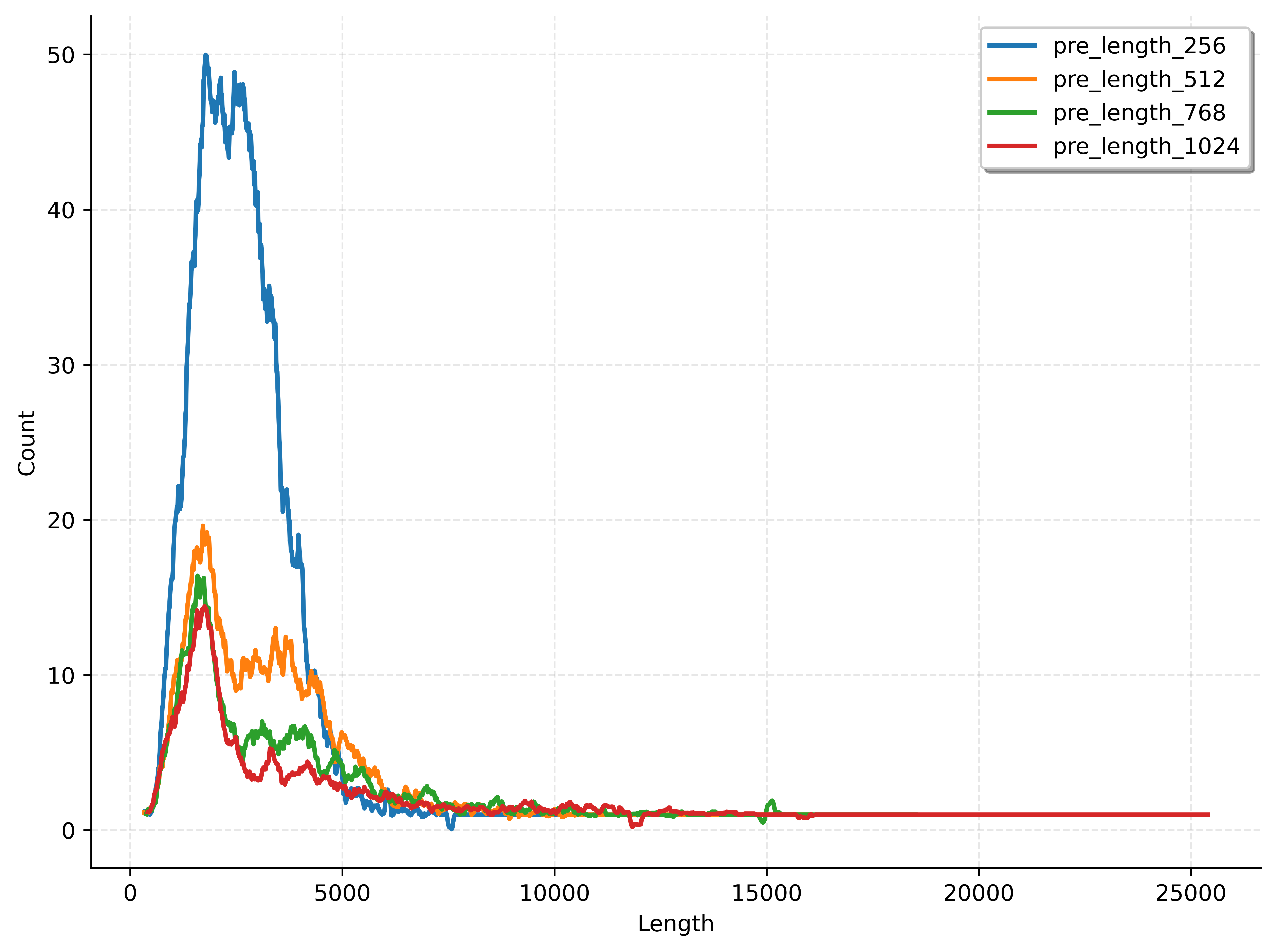}
    \caption{Character length distribution under the random selection strategy.}
    \label{fig:random}
\end{figure}

\newpage
Figure~\ref{fig:max_loss} ~\ref{fig:min_loss} ~\ref{fig:random} presents the character length distributions of answers in a medical QA dataset, comparing three sampling strategies (maximum loss, minimum loss, and random selection) across four token length constraints (256-1024). 

\newpage
\FloatBarrier 

Table~\ref{exp:ablation_metrics} presents the results of an ablation study on different sequence length settings in GraphGen, evaluating their impact on various quality metrics. The results show that increasing sequence length generally improves lexical diversity while maintaining consistent performance across evaluation metrics.

\begin{table*}[h!]
  \centering
  \resizebox{0.95\textwidth}{!}{
  \begin{tabular}{l r r r r r r r r r}
    \toprule
    \textbf{Method} & \textbf{\#Samples} & \textbf{Avg \#Tokens} & \multicolumn{6}{c}{\textbf{Results}} & \textbf{Avg Score}\\
    \cmidrule(r){4-9}
    & & & \textbf{MTLD} & \multicolumn{3}{c}{\textbf{Uni}} & \multicolumn{2}{c}{\textbf{Rew}} \\
    \cmidrule(r){5-7}
    \cmidrule(r){8-9}
    & & & & Nat & Coh & Und & Ind & Deb  \\
    \midrule
    GraphGen(256)   & 119501 & 532.6 & 71.9 & 87.8 & 95.7 & 90.4 & 84.6 & 38.0 & 74.8  \\
    GraphGen(512)   & 54287 & 657.9 & 75.8 & 87.8 & 95.7 & 90.4 & 85.0 & 31.8 & 75.2  \\
    GraphGen(768)   & 38246 & 718.1 & 75.8 & 87.9 & 95.9 & 90.4 & 84.8 & 31.0 & 75.0  \\
    GraphGen(1024)   & 32137 & 749.3 & 75.0 & 87.9 & 95.9 & 90.5 & 84.5 & 31.7 & 74.8  \\
    \bottomrule
  \end{tabular}
  } 
  \caption{\label{exp:ablation_metrics}Ablation study on quality metrics.}
\end{table*}

\newpage
\subsection{Ablation Study on Knowledge Representation Strategy}
Table~\ref{tab:ablation_agri} presents an ablation study evaluating different knowledge representation strategies for \textbf{GraphGen} on the agricultural dataset. The study compares three configurations—using only entities, only relations, and both entities and relations. Performance is measured using the ROUGE-F metric to assess the impact of different knowledge structures on model effectiveness.

\begin{table}[h!]
\centering
\begin{tabular}{l c c}
\toprule
\textbf{Method} & \textbf{Knowledge Representation Strategy} & \textbf{ROUGE-F} \\
\midrule
\multirow{3}{*}{GraphGen}  
    & Only Entities  & 50.68 \\
    & Only Relations & 51.88 \\
    & Entities + Relations & 51.77 \\
\bottomrule
\end{tabular}
\caption{Ablation study on different knowledge representation strategies for GraphGen on the agricultural dataset. The model is evaluated under three different configurations: using only entities, only relations, and both entities and relations. }
\label{tab:ablation_agri}
\end{table}

\newpage
\section{Evaluation on General and Agricultural Tasks}

Table~\ref{tab:ablation_agri_com} presents the evaluation results of different models on both general tasks and the SeedBench agricultural benchmark. The study examines six model variants, including \textbf{GraphGen}, \textbf{EntiGraph}, \textbf{Genie}, \textbf{LongForm}, \textbf{SELF-QA}, and \textbf{Wrap}. Performance is reported across multiple metrics, including GPQA, CMLU, GSM8K, BBH, MATH, Lukaemon, and various SeedBench scores.

\begin{table}[h]
\centering
\resizebox{\textwidth}{!}{
\begin{tabular}{l c c c c c c}
\toprule
\textbf{Metric} & \textbf{GraphGen} & \textbf{EntiGraph} & \textbf{Genie} & \textbf{LongForm} & \textbf{SELF-QA} & \textbf{WRAP} \\
\midrule
\multicolumn{7}{c}{\textbf{General Benchmarks}} \\
\midrule
\textbf{GPQA}    & 33.84  & 27.78  & 29.80  & 30.81  & 34.85  & 33.84  \\
\textbf{CMLU}    & 77.61  & 77.56  & 78.58  & 77.42  & 77.54  & 77.89  \\
\textbf{GSM8K}   & 80.89  & 79.45  & 80.44  & 80.74  & 80.21  & 81.20  \\
\textbf{BBH}     & 68.51  & 67.72  & 67.80  & 67.57  & 66.92  & 66.57  \\
\textbf{MATH}    & 52.45  & 52.14  & 53.73  & 52.91  & 52.99  & 54.71  \\
\textbf{Lukaemon} & 73.56  & 71.11  & 73.61  & 72.77  & 72.43  & 70.55  \\
\midrule
\multicolumn{7}{c}{\textbf{Agricultural Benchmarks (SeedBench)}} \\
\midrule
\textbf{QA-1}    & 61.75  & 65.25  & 61.25  & 61.00  & 66.25  & 63.25  \\
\textbf{QA-2}    & 75.67  & 74.49  & 75.35  & 72.59  & 75.02  & 74.09  \\
\textbf{QA-3}    & 22.46  & 30.15  & 24.69  & 24.37  & 26.43  & 26.64  \\
\textbf{QA-4}    & 50.71  & 51.61  & 51.91  & 49.48  & 49.10  & 48.76  \\
\textbf{SUM-1}   & 52.59  & 61.80  & 58.95  & 66.35  & 58.53  & 57.11  \\
\textbf{SUM-2}   & 52.20  & 63.90  & 65.30  & 71.48  & 65.44  & 63.68  \\
\textbf{RC-1}    & 96.91  & 96.91  & 96.02  & 96.46  & 96.46  & 96.02  \\
\textbf{RC-2}    & 96.67  & 87.96  & 88.63  & 90.41  & 89.09  & 90.08  \\
\textbf{RC-3}    & 77.82  & 83.66  & 83.99  & 84.45  & 83.74  & 83.97  \\
\textbf{RC-4}    & 63.34  & 71.78  & 70.95  & 72.22  & 65.44  & 66.81  \\
\textbf{RC-5}    & 75.27  & 77.60  & 74.91  & 76.35  & 75.81  & 76.17  \\
\bottomrule
\end{tabular}
}
\caption{Evaluation results of different models on general tasks and the SeedBench agricultural benchmark. General benchmarks are listed first, followed by agricultural benchmarks. In SeedBench, QA-1 corresponds to multiple-choice questions, QA-2 refers to multiple-answer questions, QA-3 involves fill-in-the-blank tasks, and QA-4 pertains to open-ended generative questions. SUM-1 represents simple summarization tasks, while SUM-2 focuses on key information extraction. RC-1 denotes multiple-choice reading comprehension, RC-2 covers multiple-answer reading comprehension, RC-3 consists of fill-in-the-blank reading comprehension tasks, RC-4 includes generative reading comprehension, and RC-5 represents subcategory classification tasks.}
\label{tab:ablation_agri_com}
\end{table}



\end{document}